\documentclass[a4paper,fleqn]{cas-sc}

\usepackage[numbers]{natbib}
\usepackage{lineno,hyperref}
\usepackage{multirow}
\usepackage{graphicx}
\usepackage{amsfonts} 
\usepackage{amssymb}
\usepackage{rotating}
\usepackage{setspace}
\usepackage{array}
\usepackage{amsmath}
\usepackage[utf8]{inputenc}
\usepackage[ruled, lined, longend, linesnumbered]{algorithm2e}
\usepackage{float}
\usepackage{soul}
\usepackage{graphicx}
\usepackage{amsmath}
\usepackage{flafter}
\def\tsc#1{\csdef{#1}{\textsc{\lowercase{#1}}\xspace}}
\tsc{WGM}
\tsc{QE}
\tsc{EP}
\tsc{PMS}
\tsc{BEC}
\tsc{DE}

\begin{document}
\let\WriteBookmarks\relax
\def\floatpagepagefraction{1}
\def\textpagefraction{.001}

\shorttitle{HumanLight: Incentivizing Ridesharing via Deep Reinforcement Learning in Traffic Signal Control}

\shortauthors{DM Vlachogiannis et~al.}

\title [mode = title]{HumanLight: Incentivizing Ridesharing via Human-centric Deep Reinforcement Learning in Traffic Signal Control}                      

%
\author[1,2]{Dimitris M. Vlachogiannis}[
                        orcid=0000-0001-5486-0274]


\ead{d.vlachogiannis@berkeley.edu}

\credit{Conceptualization, Methodology, Software, Data curation, Formal analysis, Validation, Writing - original draft, review \& editing}

\address[1]{Department of Civil and Environmental Engineering, University of California at Berkeley, Berkeley, 94720, CA, United States}

\author[3]{Hua Wei}[%
]
\credit{Methodology, Validation, Writing - review \& editing, Supervision}
\author[1,2]{Scott Moura}[%
]
\credit{Writing - review \& editing, Supervision}
\author[1,2]{Jane Macfarlane}[%
   ]
\cormark[1]
\ead{janemacfarlane@berkeley.edu}
\credit{Conceptualization, Writing - review \& editing, Supervision, Project administration}

\address[2]{Energy Technologies Area, Lawrence Berkeley National Laboratory, 1 Cyclotron Road,Berkeley, 94720, CA, United States}
\address[3]{Department of Informatics, Ying Wu College of Computing, New Jersey Institute of Technology, Suite 5700 University Heights, Newark 07102, NJ, United States}

\cortext[cor1]{Corresponding author}


\begin{abstract}
Single occupancy vehicles are the most attractive transportation alternative for many commuters, leading to increased traffic congestion and air pollution. Advancements in information technologies create opportunities for smart solutions that incentivize ridesharing and mode shift to higher occupancy vehicles (HOVs) to achieve the car lighter vision of cities. In this study, we present HumanLight, a novel decentralized adaptive traffic signal control algorithm designed to optimize people throughput at intersections. Our proposed controller is founded on reinforcement learning with the reward function embedding the transportation-inspired concept of pressure at the person-level. By rewarding HOV commuters with travel time savings for their efforts to merge into a single ride, HumanLight achieves equitable allocation of green times. Apart from adopting FRAP, a state-of-the-art (SOTA) base model, HumanLight introduces the concept of active vehicles, loosely defined as vehicles in proximity to the intersection within the action interval window. The proposed algorithm showcases significant headroom and scalability in different network configurations considering multimodal vehicle splits at various scenarios of HOV adoption. Improvements in person delays and queues range from 15\% to over 55\% compared to vehicle-level SOTA controllers. We quantify the impact of incorporating active vehicles in the formulation of our RL model for different network structures. HumanLight also enables regulation of the aggressiveness of the HOV prioritization. The impact of parameter setting on the generated phase profile is investigated as a key component of acyclic signal controllers affecting pedestrian waiting times. HumanLight's scalable, decentralized design can reshape the resolution of traffic management to be more human-centric and empower policies that incentivize ridesharing and public transit systems.

\end{abstract}


\maketitle
\section{Introduction}
Population growth along with urbanization have rendered today's transportation networks more saturated than ever before. Private vehicle ownership grows yearly, particularly since the COVID-19 outbreak \cite{Of2021}. As a result, traffic congestion has become one of the key challenges metropolitan areas are facing. Apart from increased travel times and vehicle miles traveled, the associated environmental consequences have made transportation the number one contributor to the climate crisis in America today. In many urban areas, the inconveniences of unreliable transit are high for commuters \cite{malalgoda2019transportation,diab2015bus}, which has resulted in diminishing public transit ridership, leaving private vehicles as the most attractive solution for commuters.

For policymakers, looking decades ahead, the reduction of trips performed in single occupancy vehicles (SOVs) is of paramount importance. With the COVID-19 pandemic coming to an end, incentivizing public transit use and shared modes of transportation is imperative. Ridesharing or pooling is defined as the act of two or more travellers sharing the same vehicle for a common trip. Shared rides will reduce vehicle miles traveled, energy use, and greenhouse gas emissions \cite{viegas2016shared}. Apart from the system-level benefits, the users individually benefit from shared travel costs \cite{shaheen2016casual}. Car pooling originally became popular in metropolitan regions with the establishment of High Occupancy Vehicle (HOV) lanes. People casually grouped in one vehicle to save on time, toll fares, and gas \cite{shaheen2019shared}. In the last decade, app-based pooling services that reserve, match, and process payments for rides on demand have been developed. By grouping passengers with similar origin and destination locations within a walking radius from the common route, transportation network companies in urban areas provide flexible and affordable ridesharing solutions \cite{lazarus2021pool}. Compared to the ride-alone option, users of a pooled ride are quoted a discounted price for a usually longer estimated total travel time due to the added walking time and pick up of additional passengers. 

Transit and purpose-driven shuttles (e.g. airport drop-off services) have been around for decades as fixed-guideway systems that cannot respond to dynamic passenger demand. On-demand transit-like services, typically comprised of vans, shuttles and buses, have also recently emerged and are commonly referred to as microtransit \cite{shaheen2019shared}. Particularly in the post-pandemic era which reshaped travel patterns to require greater flexibility, the on-demand services offered by microtransit with more direct routing, reduced transfers, and better coverage can significantly improve rider experience \cite{schank2022microtransit}. As opposed to traditional transit, microtransit generates routes and stops in response to real-time demand. Furthermore, microtransit vehicles are guided by GPS with real-time traffic and can therefore adapt to on-the-ground conditions such as traffic jams and road closures. Microtransit solution experiments have been mostly successful and are envisioned as a key component in transportation networks by many forward-thinking transit agencies. For those reasons, this study incorporates microtransit in the formulation of HOV adoption scenarios to represent the multimodal composition of vehicle fleets in the upcoming decades.

The dynamic travel behavior and demand fluctuations across on-demand mobility options have led to a decline of carpooling to currently only 7.79\% of the commute mode share in the United States for 2021 \cite{Of2021}. The major progress in ride-hailing matching algorithms for mobility on demand is mostly achieving cheaper rides for commuters. However, \cite{lazarus2021pool} have indicated that monetary benefits alone do not outweigh the loss in travel time poolers sacrifice to merge into a single ride. 
For pooling to be competitive, shared rides should be prioritized through smart mobility solutions that democratize travel times benefits. Currently, poolers and non-poolers share the travel time benefits of reduced vehicle counts in the transportation network thanks to poolers' shift towards HOVs.  Instead, fairer and more socially equitable mechanisms would reward poolers for their efforts to merge rides by providing reduced travel times. HOV lanes are currently the only case where poolers receive travel time benefits. In urban environments, HOV lanes are mostly reserved for freeways and highways. According to \cite{giuliano1990impact}, the introduction of HOV lanes did not lead to a significant increase in ridesharing among the population of the examined route's commuters, but only in some specific periods. Results suggested that barriers to increased pooling and mode shift are formidable, requiring the development of additional carpooling incentivizing strategies based more on travel time savings. These findings are in alignment with studies suggesting that cost reductions of pooling solutions do not outweigh the potential delays in travel time \cite{cohen2021incentivizing}. Information and connected car technologies enabling real time data sharing and high performance computation now make the window of opportunity for the implementation of smart mobility solutions greater than ever. 

In this study, we present HumanLight, a traffic management solution developed to operate with future vehicle to infrastructure (V2I) communication technologies. A novel decentralized adaptive signal control algorithm that optimizes people's throughput at intersections is formulated to support a future modal split among private vehicles, pooled rides and public transit. Our proposed solution sets the foundations for breaking free of today's dependence on cars by prioritizing people as opposed to vehicle movement. By devolving transportation policy to be more human-centric, ridesharing and public transit systems can be re-invigorated and attract the travel demand they truly merit in sustainable and multimodal urban environments. The reinforcement learning-based signal controller HumanLight is designed to reward, with more green times, vehicles carrying more people. We explore the headroom of such prioritization strategies at intersections at different rates of vehicle pooling adoption and demonstrate the higher travel time benefits for HOV adopters. 

\section{Literature Review}
\subsection{Reinforcement Learning in Traffic Signal Control}

Traffic controllers are typically classified into fixed-time, actuated, and adaptive controllers. Conventional traffic signal control (TSC) methods are mostly fixed-time and have been developed to heavily rely on pre-defined rules and assumptions on traffic conditions \cite{martinez2011survey, roess2004traffic}. Traffic splits are typically derived to alleviate traffic congestion on uniform traffic flow distribution. Webster's method \cite{koonce2008traffic} is one of the most widely-used methods in the field for a single intersection setting. It determines the optimum cycle length and phase split for a single intersection according to historical traffic data collected at different times during the day. Actuated signal controllers are more responsive to the traffic flows as they use real-time measurements from sensors. In most cases though, the timing plan parameters, such as maximum green and extension time, are optimized offline \cite{shabestary2018deep}. 

Adaptive traffic signal controllers have been shown to outperform fixed-time and actuated controllers because they can adjust traffic phase splits, or skips traffic phases, according to dynamic and unpredictable traffic demand patterns. Reinforcement Learning (RL) based TSC is a model-free and self-learning adaptive strategy. By interacting with the environment, usually a microscopic simulator, the agent learns to adapt to the evolving real-time traffic conditions \cite{wei2018intellilight, wei2019presslight, chen2020toward}. Recently, with the advent of deep learning and the use of deep neural networks to approximate key components of reinforcement learning, deep reinforcement learning (DRL) has enabled a continuous state space representation. 

A key component of the design of a RL-based TSC system is the formulation of the algorithmic setting, comprising of the state and action space and the reward function. For the state representation, different quantitative descriptions of the environment have been proposed, including queue length \cite{wei2018intellilight,bakker2010traffic,kuyer2008multiagent}, waiting time \cite{van2016coordinated,chu2019multi}, vehicle volume \cite{el2010agent,wei2019presslight} approximations of delay \cite{arel2010reinforcement}, speeds \cite{casas2017deep,el2010agent,nishi2018traffic} or phase setting \cite{aslani2017adaptive,wei2019presslight,wei2018intellilight,el2013multiagent,chen2020toward}. As for the design of the action space, two main representations are used in literature: i) the algorithm selects traffic phases in an acyclic manner \cite{shabestary2022adaptive,wei2019presslight,wei2019colight,chen2020toward,yang2019cooperative,oroojlooy2020attendlight}, ii) the algorithm, following an ordered sequence of traffic phases, determines the traffic phase splits by regulating the duration of the current phase either by keeping or switching it \cite{mannion2016experimental,wei2018intellilight,aslani2017adaptive}. The reward, or objective function of the RL problem, has been modeled with several surrogate metrics as vehicle travel time can only be measured post trip completion. Metrics that have shown to provide effective formulations of the reward include queue length \cite{wei2019colight,van2016coordinated,kuyer2008multiagent,aslani2017adaptive,salkham2010soilse}, waiting time \cite{bakker2010traffic,chu2019multi,van2016coordinated,wei2018intellilight}, speed \cite{wei2018intellilight,van2016coordinated,casas2017deep}, number of stops \cite{van2016coordinated}, throughput \cite{aslani2017adaptive,wei2018intellilight,salkham2010soilse} or even safety related features such as accident avoidance \cite{van2016coordinated,du2022safelight}.

In an effort to derive a more theoretically grounded reward function formulation and identify how much neighboring information is necessary in the state space representation, \cite{wei2019presslight} drew inspiration from max pressure (MP), a state-of-the-art (SOTA) method in adaptive traffic signal control, as proposed in \cite{varaiya2013max,lioris2016adaptive}. The MP algorithm proved that by minimizing pressure of the intersection, defined as the difference between the total queue length on incoming approaches and outgoing approaches, the risk of over-saturation is reduced and the throughput of the whole road network is maximized. \cite{wei2019presslight} tested using pressure as part of the long term reward function to achieve a formulation more theoretically justified without deploying combinations of heuristic metrics that also require weight tuning. PressLight is also the first RL model that automatically achieves coordination along arterials without prior knowledge. The framework was tested under uniform unidirectional traffic with the optimal solution being known to be the greenwave. PressLight learned to regulate consecutive traffic signals' switches with an offset equivalent to the expected vehicle travel time between intersections.

Despite directly interacting with a highly dynamic environment and learning to reach a long-term goal for traffic signal control, RL methods are susceptible to the curse of dimensionality \cite{rasheed2020deep,yau2017survey} an issue whereby the state space becomes too large. This leads to higher computational costs during the exploration of the state-action pairs resulting in a longer learning time, as well as requiring a larger storage capacity to store the learned Q-values. MPLight \cite{chen2020toward} was recently designed as a decentralized framework to address those scalability issues. The proposed approach was applied for multi-intersection control, and specifically in a large-scale network of over 1000 intersections in New York City. Thanks to the pressure-based design and the parameter sharing capabilities in model learning, MPLight showcased strong performance and generalization ability. MPLight also adopts the FRAP architecture as its base model \cite{zheng2019learning}. FRAP is a Deep Q-Learning method specifically developed for traffic signal control problems and designed based on the principles of phase competition and invariance. FRAP not only achieves superior performance with fast convergence, but is also resilient in handling complex intersection structures and multi-intersection environments.

The vast majority of SOTA studies showcase the potential of reinforcement learning in TSC in a vehicle-level optimization setting under the consideration of uni-modal traffic, neglecting the heterogeneous interests and complex interactions of multimodal traffic. Their vehicle-centric design provides optimal movement of vehicles disregarding their occupancy.

\subsection{Transit Signal Priority and Person-level Traffic Optimization}
Human mobility and optimization of people throughput at intersections is essential for establishing efficient, sustainable and equitable transportation operations. Over the last decade, automatic passenger counts (APCs), the foundation of occupancy data, are rapidly growing in their adoption within transit fleets. APCs have already been established as standard practice by several transit agencies, as real-time occupancy is a necessary piece to the complete mobility picture. The American Public Transportation Association reported in 2020 that approximately 40\% of all United States transit vehicles have APCs installed, with commuter buses exceeding 58\%. This information is already helping agencies improve transit operations by reducing bus bunching and passenger pass-ups. Monitoring passenger counts is invaluable for TSC frameworks aiming to minimize person delays, as it facilitates the implementation of traffic management strategies capable of achieving efficient and reliable public transit.

A transit signal priority algorithm was proposed in \cite{christofa2011traffic, christofa2013person}, formulated as a mixed-integer non-linear program, to minimize total passenger delay while assigning priority to transit vehicles based on their passenger occupancy. The developed macroscopic mathematical model of delay at an intersection considers both regular and transit vehicles, and assumes under-saturated traffic conditions and fixed cycle lengths and phase sequences. The work was later extended to the arterial level where intersection pairs were simultaneously optimized and schedule adherence was incorporated \cite{christofa2016arterial}. Another real-time TSC system, minimizing total person delays of cars and transit vehicles while ensuring a priority window for transit vehicles to address the issue of wasted priority due to bus arrival uncertainty, was proposed by \cite{farid2017real}. Several other systems handling transit vehicle priority in bi-modal traffic environments have been developed \cite{hu2015coordinated,yu2017person,zeng2015person,yu2018implementing}.

Only recently have traffic signal control algorithms founded on RL and accounting for person-level optimization appeared in literature \cite{rasheed2020deep}. Using available information received from high-detailed traffic sensors, \cite{shabestray2019multimodal} proposed a multimodal Deep RL based traffic signal controller that combines both regular traffic and public transit and minimizes the overall travelers’ delay through the intersection. The controller builds on the authors previous work \cite{shabestary2018deep}. The position and average speed of vehicles are used as state inputs, the change of delay is defined as the reward function, and actions are set to flexibly choose the next phase in the pre-defined phase set. \cite{long2022deep} also adopted a person-based reward function to propose an extended Dueling Double Deep Q-learning (DDDQL) algorithm, eD3QNI, to improve bus operational efficiency and handle conflicting bus priority requests. Performance is evaluated by simulation for a single intersection with two traffic demands and random arrivals, schedule deviations, and occupancies of buses. The authors also performed an exploration around the penetration rate of connected buses, illustrating that it does not affect the convergence speed but it will affect performance. Another person-based approach also restricted in an isolated intersection was proposed by \cite{wang2022human}. The reward function was built around passenger waiting time and queue length and optimized using the DDDQL network. The authors even account for non-motorized traffic including pedestrians, and report a 6.3\% decrease in waiting time compared to a baseline vehicle-level approach. To the best of our knowledge, existing literature still lacks a scalable RL-based multi-intersection traffic signal controller capable of handling multimodal traffic and generalizable to large scale networks. 

\subsection{Contributions and Organization}
The key contributions of this research work can be summarized as follows. HumanLight is the first human-centric RL-based decentralized adaptive traffic signal controller that is scalable to transportation networks of multiple intersections such as corridors and grids. HumanLight is designed to democratize urban traffic by allocating green times in a socially equitable manner. By providing travel time savings to riders of HOVs, HumanLight can be a powerful tool for policymakers to incentivize a shift from privately owned motorized transportation towards shared and equitable mobility. Our proposed algorithm's effectiveness is evaluated in a variety of road network configurations and mode share scenarios. The different multimodal scenarios include diverse distributions of SOVs, carpools, microtransit and public transportation to quantify HumanLight's potential at different levels of HOV adoption.

We introduce some key methodological novelties that enable HumanLight to achieve robust performance in person-level optimization. Inspired by the concept of pressure \cite{varaiya2013max,wei2019presslight} quantifying the degree of disequilibrium between vehicle density on the incoming and outgoing lanes, HumanLight extends the idea to person pressure minimization to achieve optimal people throughput at intersections. We also introduce the concept of active vehicles, loosely defined as those  in proximity to the intersection within the action interval window. Through systematic experiments, the utility of the active vehicle consideration during person-level optimization is shown to improve handling the high variance of vehicle occupancies in multimodal traffic environments. 

In addition to enabling reduced passenger travel times and socially equitable green time allocation at signalized intersections, HumanLight enables policymakers and traffic engineers to control the aggressiveness in the prioritization of HOVs. We achieve this via a modification in the state embedding where vehicle occupancies are encoded. This way, the travel time benefits across vehicle types of different occupancies are parameterized for the system operator to adjust rather than being generated in a fixed approach.

Experiments are conducted to establish the most effective formulation of the algorithm (state space and reward function), as well as, parameter setting and tuning. Experiments include exposure to different traffic demand scenarios and network structures while the analytics extend further from traditional traffic metrics (travel times, delays and queues) to traffic signal design aspects such as average phase durations and number of changes. To examine the potential failure points of HumanLight in network settings, the study also includes analyses on maximum vehicle queues and stopping behaviors of the different vehicle types as a result of the person-level TSC policies. 

The remainder of this paper is structured as follows. In Section \ref{sec_preliminaries}, we introduce some preliminary definitions and formulate the decentralized multi-intersection traffic signal control problem. Section \ref{sec_meth} presents the agent design and provides details on the deployed Deep Q-learning model. In Section \ref{sec_experiment}, the experimental framework is demonstrated along with the methods of comparison and evaluation metrics. Section \ref{sec_results} discusses HumanLight's performance for different network configurations and demand profiles and presents analyses justifying the algorithmic formulation and the  policymaking impacts of the proposed solution. Section \ref{sec_conclusions} concludes the study and Section \ref{future_dirs} suggests directions for future research.

\section{Problem Definition}
\label{sec_preliminaries}
\subsection{Preliminaries}
\textbf{Definition 1: Traffic Network} The traffic network is represented as a directed graph, with intersections modeled as nodes and road segments between intersections as edges. Each intersection may have both incoming and outgoing roads serving the upstream and downstream traffic respectively. Each road may be comprised of multiple lanes. Road segment $i$ of intersection $I$ is denoted by $r^I_i$ while lane $l$ of  $r^I_i$ is denoted as $r^I_{(i,l)}$. We represent the set of incoming and outgoing lanes of intersection $I$ as $L_{in}^I$ and $L_{out}^I$ respectively.

\textbf{Definition 2: Traffic Movement} Traffic movement is defined as the traffic traveling across an intersection from an incoming road to an outgoing road. We denote a traffic movement from $r^I_i$ to $r^I_j$ as $(r^I_i, r^I_j)$, in which $(r^I_i, r^I_j) = (r^I_{(i,l)}, r^I_{(j,m)})$, $l \in r^I_i \subset L_{in}^I, m \in r^I_j  \subset L_{out}^I$.

\textbf{Definition 3: Signal Phase} A traffic signal phase of intersection $I$, $\phi^I$, is defined as a permissible combination traffic movements.

\textbf{Definition 4: Movement, Phase and Intersection Pressure} Vehicle pressure of a traffic movement is defined as the difference of vehicle density between the upstream lane $r^I_{(i,l)}$ and the downstream lane $r^I_{(j,m)}$:

\begin{equation}
    \label{eq:vehicle_pressure_tm}
        p_{\nu}(r^I_{(i,l)}, r^I_{(j,m)}) = \frac{C_\nu(r^I_{(i,l)})}{C_\nu^{max}(r^I_{(i,l)})} - \frac{C_\nu(r^I_{(j,m)})}{C_\nu^{max}(r^I_{(j,m)})}
    \end{equation}

where $C_\nu(\cdot)$ is the vehicle count function, and $C_\nu^{max}(\cdot)$ the maximum permissible vehicle number for a road segment. If all lanes of the transportation network share the same maximum vehicle capacity, vehicle pressure can be simplified to capture the difference in vehicle counts between the upstream and downstream traffic.

The vehicle pressure of a traffic phase $\phi^I$ is defined as the sum of pressures over all traffic movements comprising the traffic phase:
\begin{equation}
    \label{eq:vehicle_pressure_phase}
        p_{\nu}(\phi^I) = \sum_{(r^I_{(i,l)}, r^I_{(j,m)}) \, \epsilon  \, \phi^I}^{} p_{\nu}(r^I_{(i,l)}, r^I_{(j,m)}) 
    \end{equation}

The intersection vehicle pressure is defined as the sum of pressures over all traffic movements:
\begin{equation}
    \label{eq:vehicle_pressure_inter}
        P_{\nu}^I = \sum_{(r^I_{(i,l)}, r^I_{(j,m)}) \, \epsilon  \, I}^{} p_{\nu}(r^I_i, r^I_j) 
    \end{equation}

Accordingly, we derive person intersection pressure ($P_p^I$) by substituting the vehicle count function $C_\nu(\cdot)$ of Eq. \ref{eq:vehicle_pressure_tm} with a person count function $C_p(\cdot)$:
\begin{equation}
    \label{eq:person_pressure_inter}
        P_{p}^I =  \sum_{(r^I_{(i,l)}, r^I_{(j,m)})}^{} p_{p}(r^I_i, r^I_j) 
    \end{equation}
    
In real-world world scenarios, person counts can be generated via automated passenger counters (APCs).

\begin{figure}[ht!]
\centering
\includegraphics[width=1\linewidth]{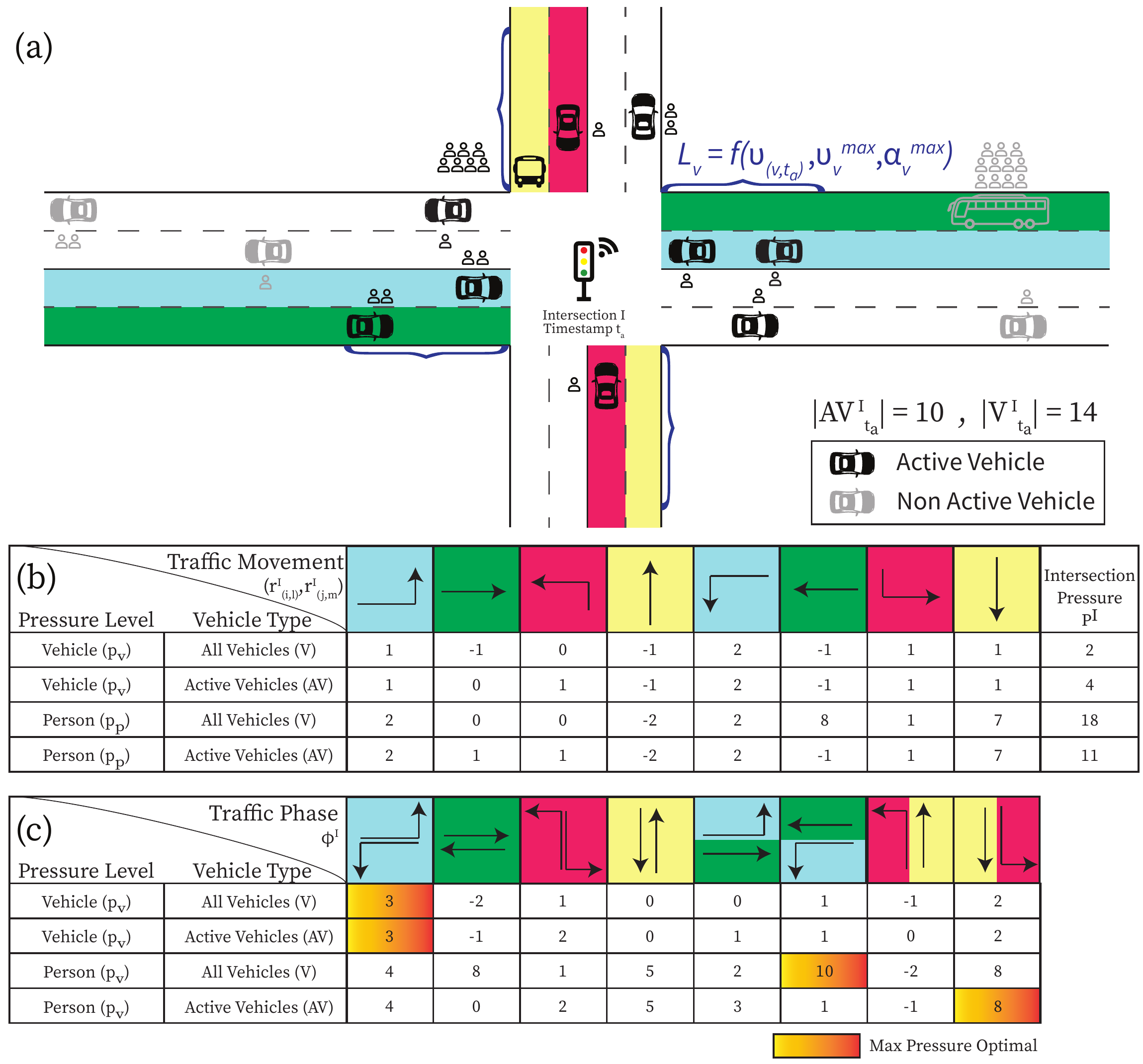}
\caption{(a) Standard four-way intersection with 8 incoming lanes. 
(b) Traffic movement illustration and pressure  calculation. (c) Traffic phase illustration and pressure aggregation.The MaxPressure optimal phase is highlighted. We emphasize two keys points: i) Accounting for people movement returns a different optimal phase than the traditional vehicle-level pressure maximization strategy. ii) The introduction of active vehicles has no impact in the vehicle-level approach (same phase is returned as optimal) but radically affects the person-level approach (the returned phases are not only different but share no common movements).}
\label{fig:AV_example}
\end{figure}

\filbreak

\begin{samepage}
\textbf{Definition 5: Active Vehicles (AV)} 

Literature has shown that vehicle-level traffic conditions can be adequately described by lane counts, including all vehicles in the incoming lanes \cite{casas2017deep,el2010agent,bakker2010traffic,wei2018intellilight}, and queue lengths, including all vehicles stopped in queue waiting to traverse the intersection \cite{chen2020toward,el2010agent,el2013multiagent,wei2019presslight}. In a person-level setting though, the distribution of people's locations along a segment may display significant variance, especially the more the world shifts to a multi-modal HOV setting. 
\end{samepage}

\filbreak
The goal is to assure that only persons capable of traversing the intersection are accounted for in our state space and reward function calculation. Inspired by the work of \cite{zhang2022expression}, we introduce the concept of active vehicles ($AV$) to describe vehicles that are in range of the controlled intersection within the action interval window. As opposed to the predecessor study, vehicle range ($L$) is not calculated under maximum speed but by deploying the position equation of motion assuming constant acceleration and a maximum vehicle speed threshold, according to Eq. \ref{eq:active_range}:

\begin{equation}
    \label{eq:active_range}
        L_{(\nu,t_{a})}(v_{(\nu, t_{a})}, v_{\nu}^{max}, a_{\nu}^{max}) = v_{\nu}^{max} \Delta t - \frac{1}{2 \; a_{\nu}^{max}}  (v_{\nu}^{max} - v_{(\nu,t_{a})})^2
    \end{equation}

where $v_{(\nu, t_{a})}$ is vehicle's $\nu$ speed at timestamp $t_a$ when action $a$ is to be taken, $v_{\nu}^{max}$,$a_{\nu}^{max}$ are the vehicle's maximum speed and acceleration, and $\Delta t$ is the action interval.

For lanes upstream of the intersection, we compute the vehicles' maximum feasible projected location at the end of the upcoming action $a$. Our concept is also extended for outgoing lanes to properly inform the algorithm regarding the impact of the applied policies. For vehicles downstream of the intersection, we are interested on whether the controller's last selected action resulted in the vehicle crossing the intersection and joining the outgoing lane. In this case, the vehicle location at the moment of the previously selected action can be either collected from past location data or approximated similarly to the incoming lane case. Eq. \ref{eq:active_vehicles} summarizes the definition of the active vehicle set for intersection $I$ at timestep $t_{a}$, $AV^I_{t_{a}}$:


\begin{equation}
    \label{eq:active_vehicles}
     AV^I_{t_{a}} = \left\{
    \nu \epsilon V |
    d^I_{(\nu,t_{a})} \leq L_{(\nu,t_{a})}  
    , \forall \nu \epsilon V 
  \right\}
\end{equation}

where $d^I_{(\nu,t_{a})}$ is the distance of vehicle $\nu$ from intersection $I$ at timestamp $t_a$, and $V$ is the set of all vehicles.

In Section \ref{sse_section}, we evaluate the efficacy of the AV consideration under different topological settings. Figure \ref{fig:AV_example} illustrates the defined concepts in a standard four-way intersection with 8 incoming and outgoing lanes. Traffic movements and phases are identified and the corresponding pressure values are calculated both at the vehicle and person-level, as well as, with or without the consideration of AV. Figure \ref{fig:AV_example} highlights that accounting for people movement returns a different optimal phase based on the MaxPressure algorithm than the traditional vehicle-level pressure maximization strategy. Another key observation is that the introduction of active vehicles has no impact in the vehicle-level approach (same phase is returned as optimal) but radically affects the person-level approach as the returned phases are not only different but share no common movements.

\subsection{Formulation: Decentralized Multi-intersection Traffic Signal Control}
Each traffic light regulated intersection in the network is controlled by a reinforcement learning agent based on the real-time people distribution along the incoming and outgoing lanes. Under the decentralized structure, agent steps are performed individually by each agent at every distinct action interval with duration $\Delta t$, while all agents of the network engage in the same learning process. Post training, the controllers will have learned to select the appropriate signal phases that maximize passenger throughput through the intersections. 

The traffic signal control problem is formulated as a Markov Decision Process $\mathcal{M} = < \mathcal{S}, \mathcal{A}, \mathcal{P}, \mathcal{R}, \gamma >$ \cite{sutton2018reinforcement}. $\mathcal{S}$ is a finite state space, $\mathcal{A}$ is a finite action space,  $\mathcal{P}(s'|s,a): \mathcal{S} \times \mathcal{A} \times \mathcal{S} \rightarrow [0,1]$ is the state transition probability from state $s$ to state $s'$ determined by action $a$, and $\mathcal{R}(s,a): \mathcal{S} \times \mathcal{A} \rightarrow \mathbb{R}$ is the reward function defined as $\mathcal{R}(s,a) =  \mathbb{E}[R_{t+1}|s_t = s, a_t = a]$. At timestep $t$, agent $I$ aims to learn policy $\pi(a_t = a|s_t= s)$ returning the optimal action $a$ given the state $s$ to maximize the discounted reward:

\begin{equation}
    \label{eq:discounted_reward}
    \mathcal{J}_t(\pi)= \sum\limits_{m=0}^{\infty} \gamma ^m R_{t+m+1}
\end{equation}

The agent calibrates its estimates of the executed action’s utility based on environmental feedback and will potentially adjust the rates of the actions leading up to the current action.

\section{Methodology}
\label{sec_meth}
\subsection{Agent Design}
In this section, we describe the basic constructs of the reinforcement learning algorithm. The state and action spaces and the reward function are defined at the level of a single intersection $I$.
\begin{itemize}
    \item State space: The state space includes the number of persons drawn from the active vehicles in each incoming lane and in each outgoing lane of the intersection, $C_p^{AV}(r^I_{(i,l)})$ and $C_p^{AV}(r^I_{(i,m)})$ respectively with $l \epsilon L_{in}^I$ and $m \epsilon L_{out}^I$. The state space also includes the current traffic signal phase $\phi^I$.
    \item Action space: The traffic phase $\phi^I_{t+1}$ for the next action interval. For most common transportation intersections with up to four entering and existing roads, the maximum compatible and non-conflicting phase combinations are eight for each isolated intersection (Fig. \ref{fig:AV_example}). As per common practice in RL research on traffic signal control \cite{wei2019colight,wei2018intellilight,wei2019presslight,wei2019survey,chen2020toward}, we adopt the acyclic phase paradigm which, as opposed to the cyclic paradigm, does not require a predetermined phase sequence to be imposed and for all available phases to appear at least once within a cycle. By enabling phase shortening and skipping, acyclic phasing schemes have demonstrated superior performance \cite{kanis2021back} and are already implemented in urban environments (e.g. Amsterdam). However, our RL representation is still fully transferable to a cyclic phase modelling scheme.

    \item Reward function: For each individual agent, the reward is defined as the opposite of person intersection pressure as derived from Eq. \ref{eq:person_pressure_inter}:
    \begin{equation}
        \label{eq:reward_function}
         R^I = {-} P^I_p
    \end{equation}

\end{itemize}
\subsection{Base Model: FRAP}
The FRAP architecture \cite{zheng2019learning} is adopted as the base model for our traffic signal control system. FRAP is a Deep Q-Learning method specifically designed for traffic signal control problems. In vehicle-level traffic optimization, FRAP captures the competition relation between different traffic movements achieving superior performance and improved convergence. The authors also illustrated the model's transferability and adaptability to different traffic signal settings, roads structures and unbalanced traffic flows.   \begin{samepage}
The model is designed based on the principals of: 
\begin{itemize}
    \item Competition: Phases with higher traffic demand need to be prioritized.
    \item Invariance: Flipping or rotation of traffic flows along the intersection should not affect algorithmic performance. 
\end{itemize}
\end{samepage}

The prediction of the Q-values is divided into three stages:
\begin{enumerate}
    \item Phase demand modeling: Obtains a representation for the demand of each signal phase. The state features for each movement are extracted from the simulator and passed through two fully-connected layers. The outputs for the non-conflicting movements of every phase are added to generate the phase demand for green signal.  
    \item Phase pair representation: Establishes the phase pairs demand embeddings and applies convolutional layers with 1x1 filters to extract the phase pair representations, enabling the competing phases to interact with each other.
    \item Phase pair competition: Predicts Q-values for each phase. The relative priorities of each phase are computed by multiplying the phase pair demand representation with a phase competition mask and applying an additional convolutional layer with 1x1 filter. The row sum of this pairwise competition matrix returns the array of Q-values, whose maximum value dictates the action to be selected.
\end{enumerate}

All agents (traffic signal regulated intersections) of the evaluated network share the same FRAP model. The replay memory stores experiences (observations, actions and rewards) from all intersections which are used for the model's parameter updating.

\begin{table}[ht!]
\caption{Simulation and model parameters.}
\label{tab:sim_model_params}
\begin{tabular}{lclcl}
\hline
\textbf{Simulation} & \textbf{}  &\\ \hline
Simulation time & 1 hour &  CityFlow step length & 1 s/step & \\
Number of phases & 8 & Type of Signal & Acyclic & \\
Yellow and all red period & 5 s  & Action interval & 10 s & \\ 
Vehicle maximum speed & 40km/hr & Vehicle maximum acceleration & 2 m/s$^2$ & \\
Vehicle maximum deceleration & 4.5 m/s$^2$ & Vehicle minimum gap & 2.5 m & \\ 
Headway & 2 s & SOV and Carpool vehicle length & 5 m \\
Microtransit (occupancy 10) vehicle length & 7 m & Microtransit (occupancy 20) vehicle length & 9.5 m & \\ 
Microtransit (occupancy 30) vehicle length & 12 m & Public transit vehicle length & 15 m \\\hline
\textbf{FRAP Model} &  &  \\ \hline
Discount factor $\gamma$ & 0.6 & Episodes & 200 &   \\
Learning rate & 0.001 &  Buffer size & 10000 &  \\
Batch size & 32 &  Greedy policy $\epsilon$ & 1 $\rightarrow$ 0 &  \\
Learning start & 0 & Optimizer & Adam & \\
Number of layers (phase demand modeling) & 2 & Number of convolution layers w/ 1x1 filters & 20 &  \\\hline 
\end{tabular}
\end{table}

\section{Experiment}
\label{sec_experiment}
All of our experiments are conducted in an open-source microscopic traffic simulator called CityFlow \cite{zhang2019cityflow}, whose interface supports RL training for large-scale traffic signal control. Each green signal is followed by three-second yellow time during which -if possible- vehicles always choose to stop, and two-second all red time to prepare the signal phase transition. Right turning vehicles are considered to be allowed to turn during any phase but always yield priority to through moving vehicles in case of conflict. The action interval  set to $10s$ as per common practice in RL research on acyclic TSC \cite{wei2018intellilight, wei2019colight,wei2019presslight,wei2019survey,chen2020toward}. \cite{wei2018intellilight} also concluded that the action time interval has minimal influence on performance when in the range of 5 to 25 seconds. The adopted values for all simulation and model parameters are detailed in Table \ref{tab:sim_model_params}. 

\begin{figure}[ht!]
\centering
\includegraphics[width=1.0\linewidth]{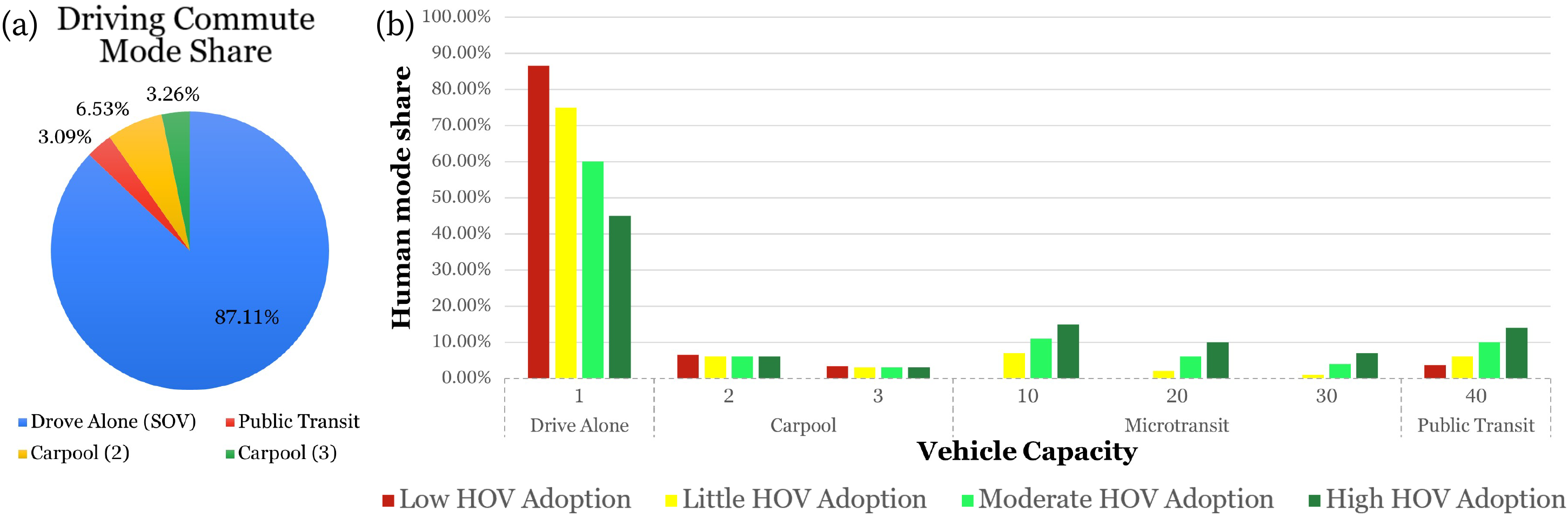}
\caption{(a) Adjusted driving commute mode share statistics drawn from the Bureau of Transportation Statistics for the year 2021. (b) Vehicle occupancy profiles for the HOV adoption scenarios (low, light, medium and high). Across each of the four scenarios, 15\% of the single occupancy vehicle individuals shift to microtransit vehicles and public transit. Carpool ridership has been preserved at the same levels.}
\label{fig:mode_shift}
\end{figure}

\subsection{Mode Share Scenarios}
Our experiments are set up by establishing a total people count served by the transportation network when it is operating almost at capacity. This is assessed through a critical movement analysis on the intersections' hourly expected flows \cite{urbanik2015signal}. We maintain those people counts constant throughout the mode share scenarios investigating different levels of HOV adoption. Note that we assume no latent demand in the network due to the increased level of service in order to quantify the benefits of mode shift with person-based RL traffic signal controlled intersections.

Determining mode share percentages is a complex task and distributions may vary per location, especially after the COVID-19 pandemic. For the lowest HOV adoption scenario, the selected numbers aim to reflect the significant reduction transit ridership took globally during the pandemic. Even, in areas such as the Bay Area where transit was highly adopted, recent studies estimate that between 66\% and 78\% of Bay Area commuters drive alone when commuting \cite{Cowan2021}. As shown in Fig. \ref{fig:mode_shift}a, commute mode data was drawn from the Bureau of Transportation Statistics for 2021 in California to extract the mode share for persons driving vehicles. People walking, working from home and biking were excluded. The percentage of the taxi, motorcycle, or other class was merged into the drive alone class. The carpooling class is split to two sub-classes of occupancy two and three with ratios 2/3 and 1/3 respectively. Fig. \ref{fig:mode_shift}b summarizes the defined mode share scenarios. For our experimental setup, commute mode changes across scenarios with approximately 15\% of the single occupancy vehicles individuals shifting to microtransit vehicles and public transit, while carpool ridership has been preserved at the same levels. The low HOV adoption scenario has transit set to its lowest levels of the last decade during allowing us to explore the potential of person-level traffic signal control optimization in a wide spectrum of HOV penetration environments.

\subsection{Infrastructure Configurations}
\begin{figure}[ht!]
\centering
\includegraphics[width=.90\linewidth]{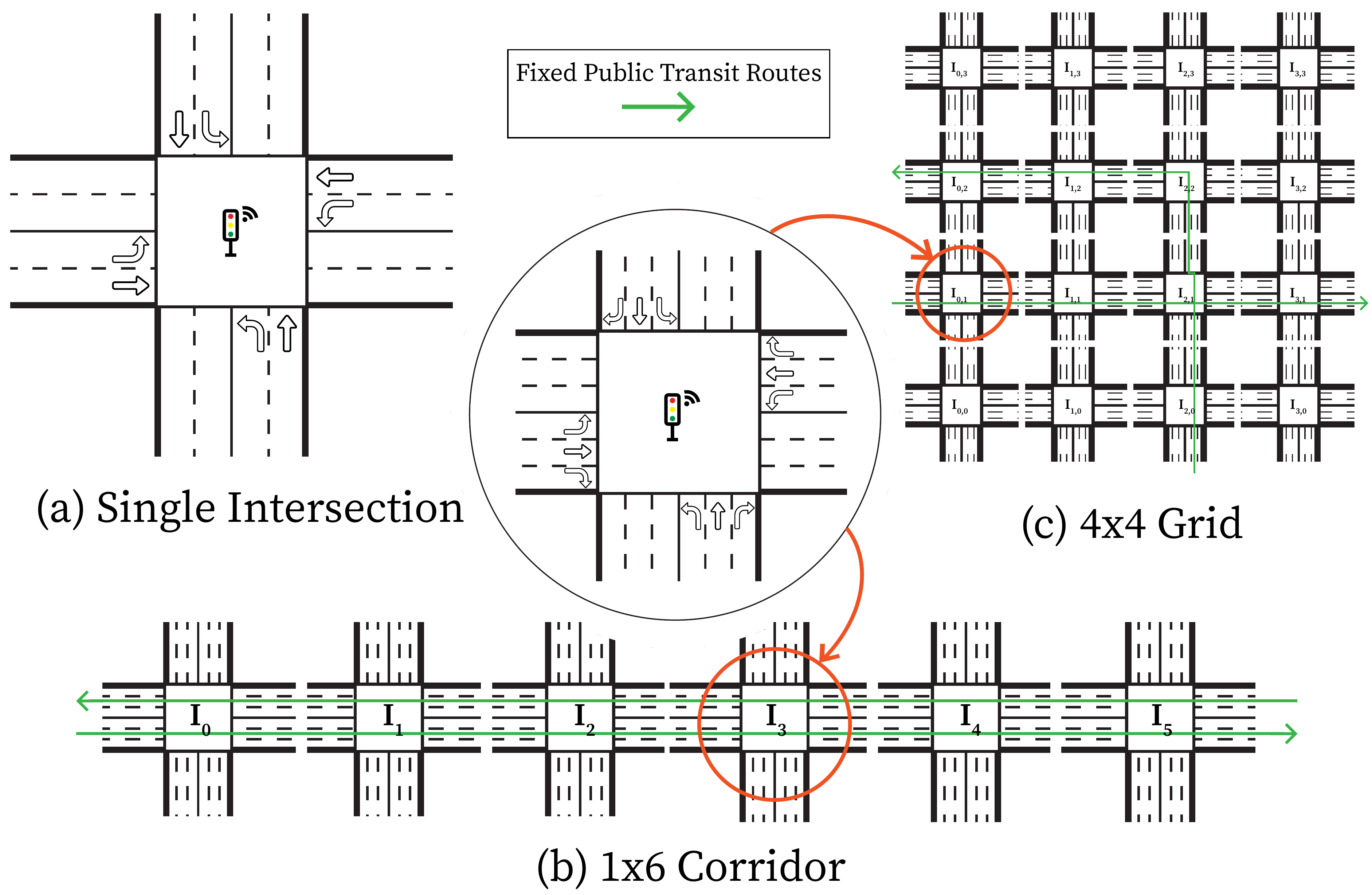}
\caption{Road Network Configurations: (a) Isolated four-way intersection where each road segment has one through and one left turning lane, (b) Six intersection corridor and (c) 4X4 grid of intersections with three lanes per road segment (one through, one left turning, and one right turning).  Fixed public transit routelines are illustrated in green.}
\label{fig:roadnets}
\end{figure}

The  set of synthetic experiments were designed to test the performance and efficiency of HumanLight at different road network configurations. These are outlined below and displayed in Fig. \ref{fig:roadnets}:

\begin{enumerate}
    \item Single intersection
    \begin{table}[ht!]
\caption{Average flow per direction per lane in the single intersection experiment across all demand scenarios.}
\label{tab:demand_single_inter}
\begin{tabular}{lccccc}
                                                              & \multicolumn{1}{l}{}                & \multicolumn{1}{l}{}             & \multicolumn{1}{l}{}               & \multicolumn{1}{l}{}            & \multicolumn{1}{l}{}                                               \\ \cline{2-6} 
\multicolumn{1}{c|}{}                                         & \multicolumn{2}{c|}{\cellcolor[HTML]{EFEFEF}North/South (veh/hr/lane)} & \multicolumn{2}{c|}{\cellcolor[HTML]{EFEFEF}East/West (veh/hr/lane)} & \multicolumn{1}{c|}{\cellcolor[HTML]{EFEFEF}}                      \\ \cline{1-5}
\multicolumn{1}{|l|}{\cellcolor[HTML]{EFEFEF}\textbf{Demand Scenario}} & \multicolumn{1}{c|}{Through}        & \multicolumn{1}{c|}{Left}        & \multicolumn{1}{c|}{Through}       & \multicolumn{1}{c|}{Left}       & \multicolumn{1}{c|}{\multirow{-2}{*}{\cellcolor[HTML]{EFEFEF}v/c}} \\ \hline
\multicolumn{1}{|l|}{Low HOV adoption}                                & \multicolumn{1}{c|}{220}            & \multicolumn{1}{c|}{220}         & \multicolumn{1}{c|}{350}           & \multicolumn{1}{c|}{220}        & \multicolumn{1}{c|}{0.99}                                          \\ \hline
\multicolumn{1}{|l|}{Light HOV Adoption}                            & \multicolumn{1}{c|}{194}            & \multicolumn{1}{c|}{194}         & \multicolumn{1}{c|}{307}           & \multicolumn{1}{c|}{194}        & \multicolumn{1}{c|}{0.87}                                          \\ \hline
\multicolumn{1}{|l|}{Moderate HOV Adoption}                         & \multicolumn{1}{c|}{157}            & \multicolumn{1}{c|}{157}         & \multicolumn{1}{c|}{249}           & \multicolumn{1}{c|}{157}        & \multicolumn{1}{c|}{0.71}                                          \\ \hline
\multicolumn{1}{|l|}{High HOV Adoption}                           & \multicolumn{1}{c|}{121}            & \multicolumn{1}{c|}{121}         & \multicolumn{1}{c|}{190}           & \multicolumn{1}{c|}{121}        & \multicolumn{1}{c|}{0.54}                                          \\ \hline
\end{tabular}
\end{table}
    \item Corridor of six intersections (1x6) with buses operating only through the corridor.
    \item Grid of intersections (4x4) with two fixed bus route-lines.
\end{enumerate}

For the single four-way intersection experiment, each road segment has one through and one left turning lane while average hourly traffic demands are described in Table \ref{tab:demand_single_inter}. We impose high left turning volumes equal to the through-moving traffic of the secondary direction. This allows us test the robustness of HumanLight on handling demand of different distributions across directions compared to the following configurations. For the corridor and grid configurations, each road segment has one through, one left turning and one right turning lane. The turning ratios at intersections are on average set as 10\% left, 60\% straight and 30\% right as statistical analyses on real-world data sets have showcased \cite{chen2020toward,VLACHOGIANNIS2023104112}. The road networks have segments of 300m in length and connect at four-way intersections. The synthetic data fed into the simulator include bi-directional flows with turning traffic. To assess our traffic signal control models under various  traffic demands the configurations of Table \ref{tab:demand_grid} are realised describing the different mode shift scenarios. The maximum speeds  are set equally for all vehicle types since the imposed values for speed limits are not that high at 40 $km/hr$.

\begin{table}[ht!]
\caption{Demand configurations for corridor (1x6) and grid (4x4) experiments.}
\label{tab:demand_grid}
\begin{tabular}{|l|c|}
\hline
\rowcolor[HTML]{EFEFEF} 
\textbf{Demand Scenario} & \textbf{\begin{tabular}[c]{@{}c@{}}Arrival Rate\\ (veh/hr/road)\end{tabular}} \\ \hline
Low HOV adoption                 & 700                                                                           \\ \hline
Light HOV adoption             & 571                                                                              \\ \hline
Moderate HOV adoption             & 469                                                                              \\ \hline
High HOV adoption            & 372                                                                              \\ \hline
\end{tabular}
\end{table}

\subsection{Methods of comparison}
In this section, we introduce classical TSC methods in the transportation field \cite{roess2004traffic,wei2019survey,martinez2011survey}, and current RL-based methods. For the purposes of evaluating HumanLight's performance, all methods including parameter setting are tuned. The best performing method will be used as benchamrk for the computation of the evaluation metrics.
 \begin{itemize}
    \item{\textbf{Fixed Time Controller - Webster's Formula} \cite{koonce2008traffic}}: Fixed-time controllers use a predetermined cycle and phase time plan. Webster's method calculates the cycle length and phase split for a single intersection setting. Under the assumption of uniform traffic flow during a certain period of time, using the closed-form solution of Eq. \ref{eq:websters_formula_cycle_length}, Webster's formula derives the optimal cycle length $C$, minimizing the travel time of all vehicles passing the intersection.

    \begin{equation}
    \label{eq:websters_formula_cycle_length}
        C(V_c) = \frac{N*t_L}{1-\frac{V_c}{3600/h*PHF*(v/c)}}
    \end{equation}
    
    where $N$ represents the number of phases, $t_L$ the total loss time per phase (used to model the all-red time and the acceleration and deceleration of vehicles), parameter $h$ the saturation headway time, $PHF$ the peak hour factor (used to model traffic demand fluctuations within peak hours), and $v/c$ is desired volume-to-capacity ratio. $V_c$ expresses the sum of all critical lane volumes, with $V_c = \sum_{p_i}^{N}V_c^{p_i}$, where $V_c^{p_i}$ is the critical lane volume for phase $p_i$. The critical lane volume is determined from the approaching lane with the highest ratio of traffic flow to saturation flow during a phase.

    As for the phase split, having established the cycle length, green times are calculated proportionally to the critical lane volumes served by each phase according to Eq. \ref{eq:websters_formula_phase_split}:
    \begin{equation}
    \label{eq:websters_formula_phase_split}
        \frac{t_{p_i}}{t_{p_j}} = \frac{V_c^{p_i}}{V_c^{p_j}}
    \end{equation}
    where $t_{p_i}$ and $t_{p_j}$ correspond to the phase duration for phases $p_i$ and $p_j$ respectively.



    
    \item \textbf{Self-Organizing Traffic Light (SOTL)} \cite{gershenson2004self,cools2013self}: SOTL is an actuated method that adaptively regulates traffic lights based on a hand-tuned threshold on the number of waiting vehicles. The controller switches the phase of a lane to green if the required minimum green phase duration is met for the current phase and provided that the number of vehicles on the lane exceeds the hand-tuned threshold. SOTL resembles a fully-actuated controller but instead of sending requests for green when a single vehicle is approaching, it does so when the number of vehicles exceeds the threshold.

    \item \textbf{MaxPressure} \cite{varaiya2013max}: Max Pressure sets the optimization objective as minimizing the vehicle pressure of phases for individual intersections, as defined in Section \ref{sec_preliminaries}. The method greedily selects the phase with the maximum pressure, activates it and keeps the selected phase for a given period of time $t_{min}$. A sensitivity analysis on parameter $t_{min}$ is carried out every time the algorithm is applied to identify the best performing value.

    \item \textbf{MPLight} \cite{chen2020toward}: MPLight is a state-of-the-art deep reinforcement learning method enabling large scale road network control. Aside from parameter sharing, it accommodates a SOTA Q-network model structure (FRAP) as proposed in \cite{zheng2019learning} and can incorporate the concept of vehicle pressure in the reward function as proposed in \cite{wei2019presslight}.
 \end{itemize}

\subsection{Evaluation Metrics}
Both person and vehicle based metrics will be used to evaluate the impact of the proposed traffic signal control strategy. These are:

\begin{enumerate}
    \item Average Vehicle Travel Time (AVTT) \& Average Person Travel Time (APTT)
    \begin{equation}
        \label{eq:AVTT}
             AVTT = \sum\limits_{\nu \epsilon V} TT_{\nu} \bigg/ |V|
        \end{equation}
    where $TT_{\nu}$ denotes the travel time of vehicle $\nu$ and $|V|$ is the cardinality of the vehicle set $V$, and
    \begin{equation}
        \label{eq:APTT}
             APTT = \sum\limits_{\nu \epsilon V} TT_{\nu} \cdot O_{\nu} \bigg/ \sum\limits_{\nu \epsilon V} O_{\nu} 
        \end{equation}
    where $O_{\nu}$ is the occupancy of vehicle $\nu$.

    \item Vehicle Queue Length (VQL) \&  Person Queue Length (PQL)
    \begin{equation}
        \label{eq:AVQL}
             VQL(t) = \bigg|\{\nu \epsilon V | v_{(\nu, t)} \leq v_{stop}, \forall \nu \epsilon V \}\bigg|
        \end{equation}   
    where $v_{stop}$ is the speed lower bound for vehicles to be considered in motion and is set at 0.1 $m/s$, and 

    \begin{equation}
        \label{eq:APQL}
             PQL(t) = \sum\limits_{\nu \epsilon V|v_{(\nu, t)} \leq v_{stop}} O_{\nu} 
        \end{equation}   
    
    \item Average Vehicle Delay (AVD) \& Average Person Delay (APD)
    \begin{equation}
        \label{eq:AVD}
             AVD = \sum\limits_{\nu \epsilon V} (TT_{\nu} - FFTT_{\nu}) \bigg/ |V|
        \end{equation}   
    where $FFTT_{\nu}$ is the free- flow travel time of vehicle $\nu$, derived accounting for the maximum speed allowed on each road segment along the vehicle's route
    , and

    \begin{equation}
        \label{eq:APD}
             APD = \sum\limits_{\nu \epsilon V} (TT_{\nu} - FFTT_{\nu}) \cdot O_{\nu} \bigg/ \sum\limits_{\nu \epsilon V} O_{\nu} 
        \end{equation}   
\end{enumerate}

\begin{figure}[ht!]
\centering
\includegraphics[width=1.0\linewidth]{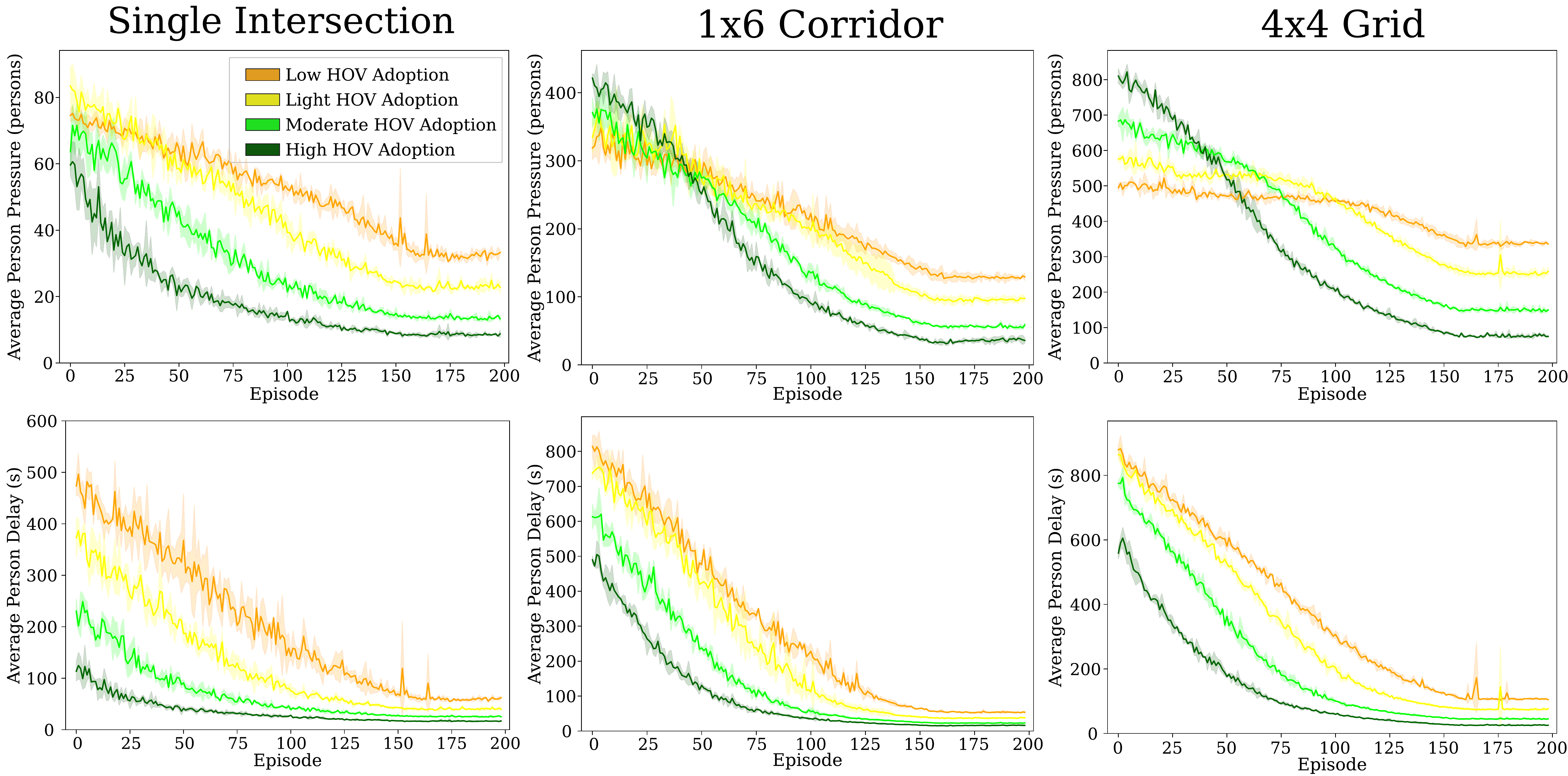}
\caption{Convergence curves of HumanLight's reward function (top) and average person delay (bottom) throughout the learning episodes for all network structures across all HOV adoption scenarios: Episodic evolution of average person pressure is approximating average person delays. Average person queues display similar trends and are provided in Fig. A\ref{fig:person_queue_convergence}.}
\label{fig:Pressure_persondelay_convergence}
\end{figure}

\section{Results}
\label{sec_results}
This section quantitatively highlights the key contributions of HumanLight. Firstly, the overall performance of the algorithm is discussed compared to SOTA controllers. Then, the socially equitable allocation of green times achieved by HumanLight is illustrated. A vehicle stopping behavior analysis as well as edge cases of vehicle queues are displayed to assure the stability of the framework. The impact of incorporating the concept of active vehicles in the formulation of our RL model is quantified through systematic experiments at different network structures. The capability of HumanLight to regulate the aggressiveness of HOV prioritization is showcased via a modification in the state embedding. Finally, the impact of the discount factor parameter $\gamma$ on the generated phase profile is investigated.

The presented results have been reproduced and validated across all network configurations considered (single intersection, corridor and grid). We selectively include results from specific networks in this section, while adding the rest to the Appendix to avoid overwhelming readers with potentially similar patterns. As per common practice in RL research \cite{long2022deep,henderson2018deep, vlachogiannis2020reinforcement,dietterich1998maxq}, multiple, in our study three, independent trials of learning are run for each scenario and the evaluation metrics (travel times, delays, queues) reported in the results are derived from averaging the last 20 episodes across all runs to smooth out episodic oscillations. The tables both in the main body and in the appendix section provide the standard deviations of those values in parenthesis.

\begin{figure}[ht!]
\centering
\includegraphics[width=1.0\linewidth]{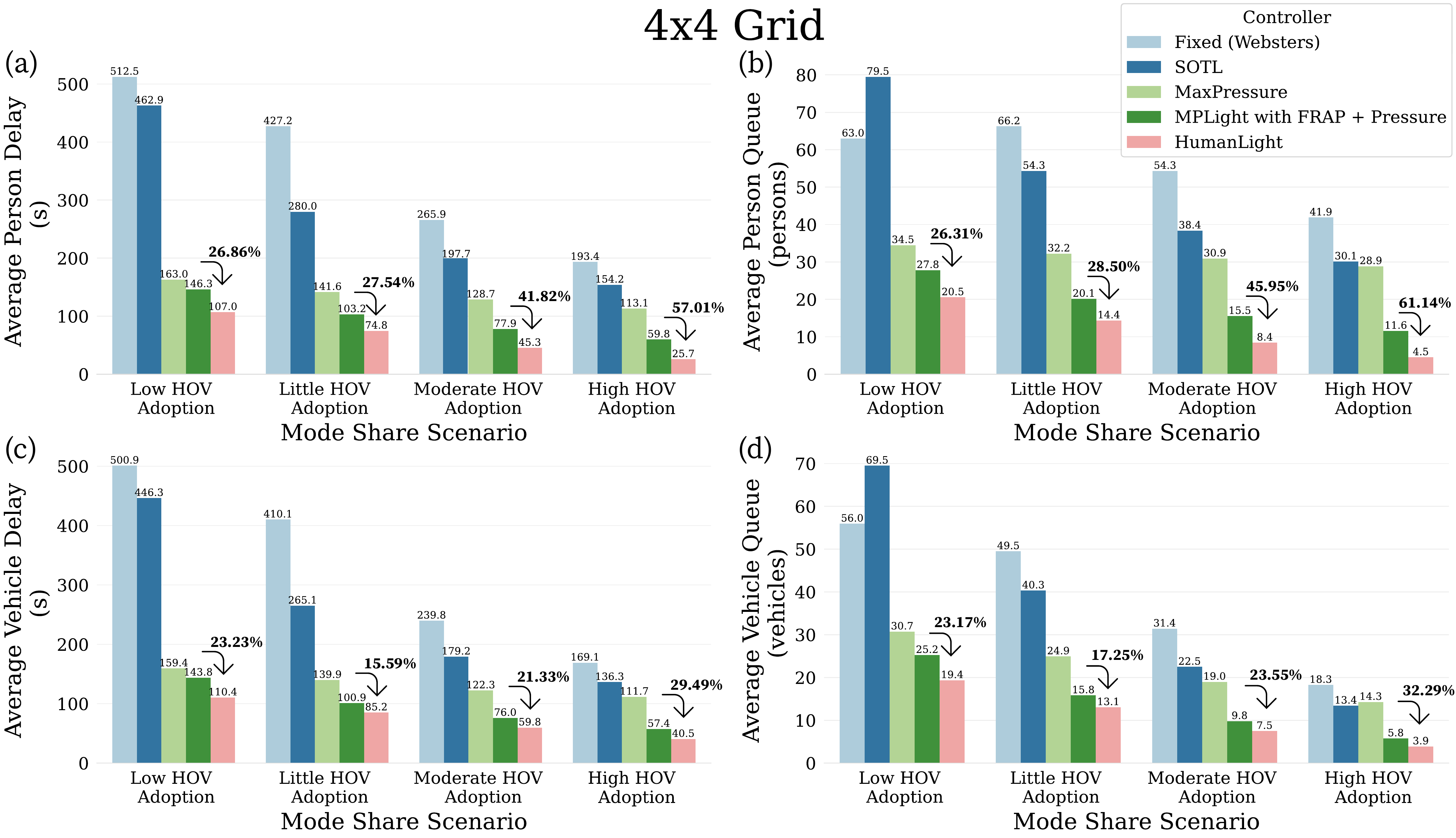}
\caption{Performance evaluation of HumanLight versus the SOTA vehicle-level optimization controllers in the 4x4 grid setting: the person-level benefits (a\&b) of HumanLight increase as more people shift towards HOV alternatives. Thanks to the consideration of active vehicles, vehicle-level metrics (c\&d) also improve. The corresponding results for the single intersection and corridor experiments are provided in Tables  A\ref{tab:si_metrics}, A\ref{tab:1x6_metrics} and A\ref{tab:4x4_metrics}.}
\label{fig:4x4_vehiclevsperson_all}
\end{figure}
\subsection{Convergence and Overall Performance}
HumanLight displays strong performance across all considered road network configurations. The algorithm consistently converges after approximately 160 episodes of training across all HOV adoption scenarios (Fig. \ref{fig:Pressure_persondelay_convergence} top). We observe that average person intersection pressure, the opposite of our reward function, follows a similar diminishing trend as average person delays (Fig. \ref{fig:Pressure_persondelay_convergence} bottom) and queues (Fig. A\ref{fig:person_queue_convergence}), validating the suitability of person pressure for the task of optimizing people throughput at intersections.

Fig. \ref{fig:4x4_vehiclevsperson_all} summarizes the performance of the various controllers in a 4x4 grid setting. Not distinguishing between vehicle types when it comes to determining intersection priorities, vehicle-level optimization controllers render similar average vehicle and person delays. We observe however, that vehicle delays are consistently lower than person delays in the corridor (Table A\ref{tab:1x6_metrics}) and grid experiments (Fig. \ref{fig:4x4_vehiclevsperson_all}, Table A\ref{tab:4x4_metrics}). This is attributed to transit's routelines never including a right turn in those configurations (Fig. \ref{fig:roadnets}). With the most dense, occupancy-wise, vehicle type never making right turns which are independent from the current phase, person travel times are expected to be slightly higher. Instead, in the single intersection setting where no right turns exist, the trend does not appear (Table A\ref{tab:si_metrics}).

By rewarding HOV riders with more green times, HumanLight achieves reduced person delays and queues. The higher the penetration of HOVs becomes, the improvements from the SOTA vehicle-level optimization controller in person delays and queues increase, even exceeding 55\% in the high HOV adoption scenario. In Fig. \ref{fig:4x4_vehiclevsperson_all}c\&d, we observe that HumanLight achieves improvements even on vehicle-level metrics, attributed to the consideration of active vehicles in the RL problem formulation. 

\subsection{Socially Equitable Allocation of Green Times}

\begin{figure}[ht!]
\centering
\includegraphics[width=1.0\linewidth]{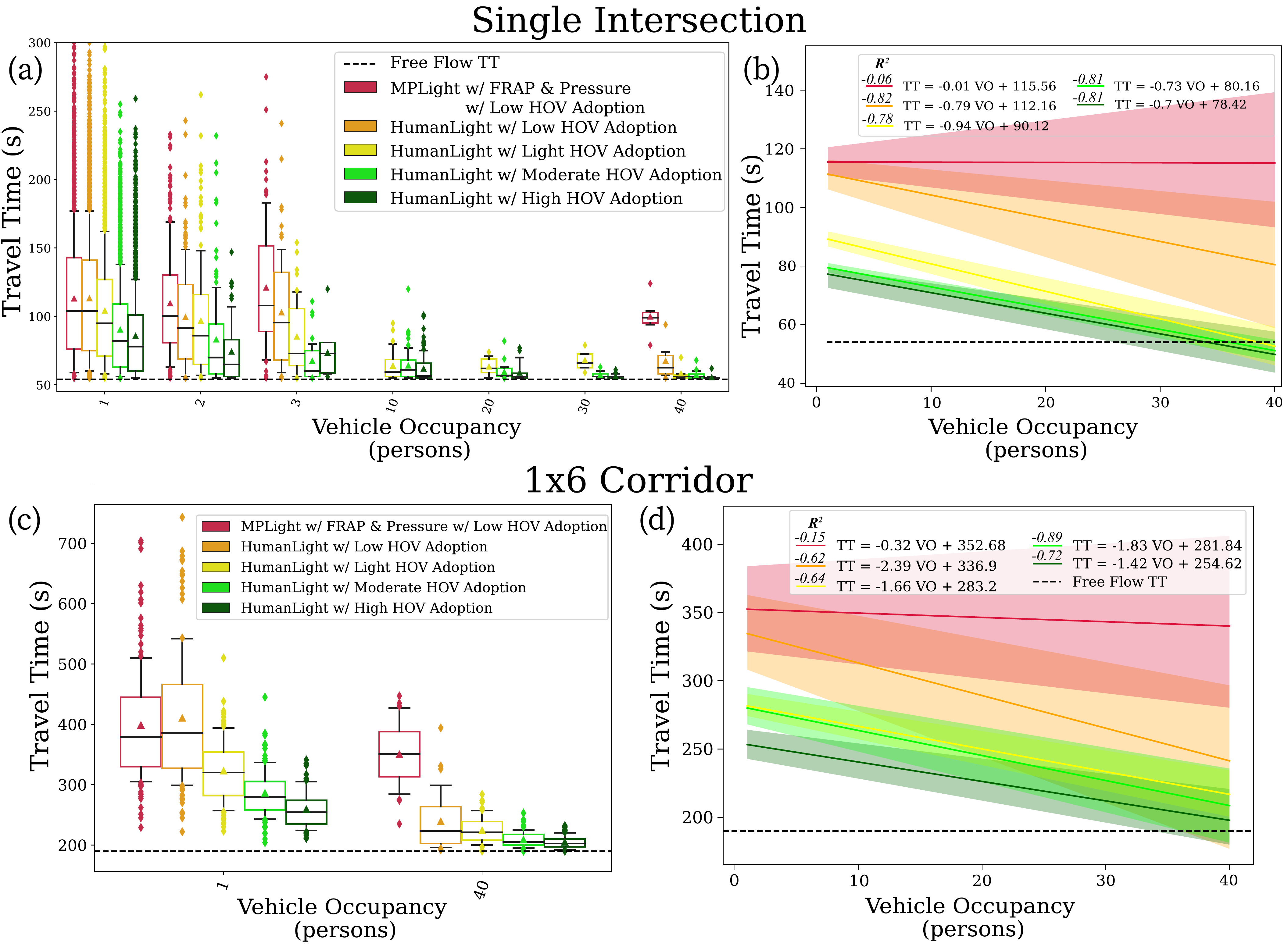}
\caption{(a,c) Travel time distribution for different vehicle types over the last episode (200). (b,d) HumanLight prioritizes vehicles based on their occupancy as shown by the linear least-squares regression fitted on the box-plot means of all vehicle types over the last 20 episodes of each run.}
\label{fig:SI_1x6_TTvsOCC}
\end{figure}

Apart from superior performance compared to SOTA approaches, HumanLight achieves a fair allocation of green times to vehicles based on the number of passengers they are carrying. Figure \ref{fig:SI_1x6_TTvsOCC} demonstrates the distributions of travel times for the different vehicle types over the last episode of training (left) and the linear least-squares regression fitted on the box-plot means over the last 20 episodes of each run to smooth out the effect of fluctuations across episodes (right). All vehicles are considered in the single intersection setting, while only vehicles traversing all six intersections are accounted for in the corridor experiments to assure all vehicles share the same route length and thus have comparable travel time. In Fig. \ref{fig:SI_1x6_TTvsOCC}c, carpools and microtransit are purposefully not displayed due to data sparsity, as limited amount of some of those vehicles fulfilled the entire six intersection route especially in the lower HOV adoption scenarios. 

Examining those distributions of travel times, the following key observations are derived. Firstly, the vehicle-level SOTA optimization controller (MPLight w/ FRAP \& Pressure, shown in red) provides similar travel times across all vehicle types. Instead, HumanLight prioritizes vehicles based on their occupancy with the differences in travel time being eminent even at the low HOV adoption scenario (orange) with a strong coefficient of determination $R^2$ between the two variables. This trend being validated at the corridor level illustrates the scalability of the model to larger networks and paves the way for large-scale evaluations with a focus on route-level effects for public transit. The lower bound of the minimum travel time is only achievable in an unrealistic scenario where all vehicles arrive at the intersection from directions served by non conflicting maneuvers within the action interval window.

\subsection{Vehicle Stopping Behavior and Queues}
\begin{figure}[ht!]
\centering
\includegraphics[width=0.9\linewidth]{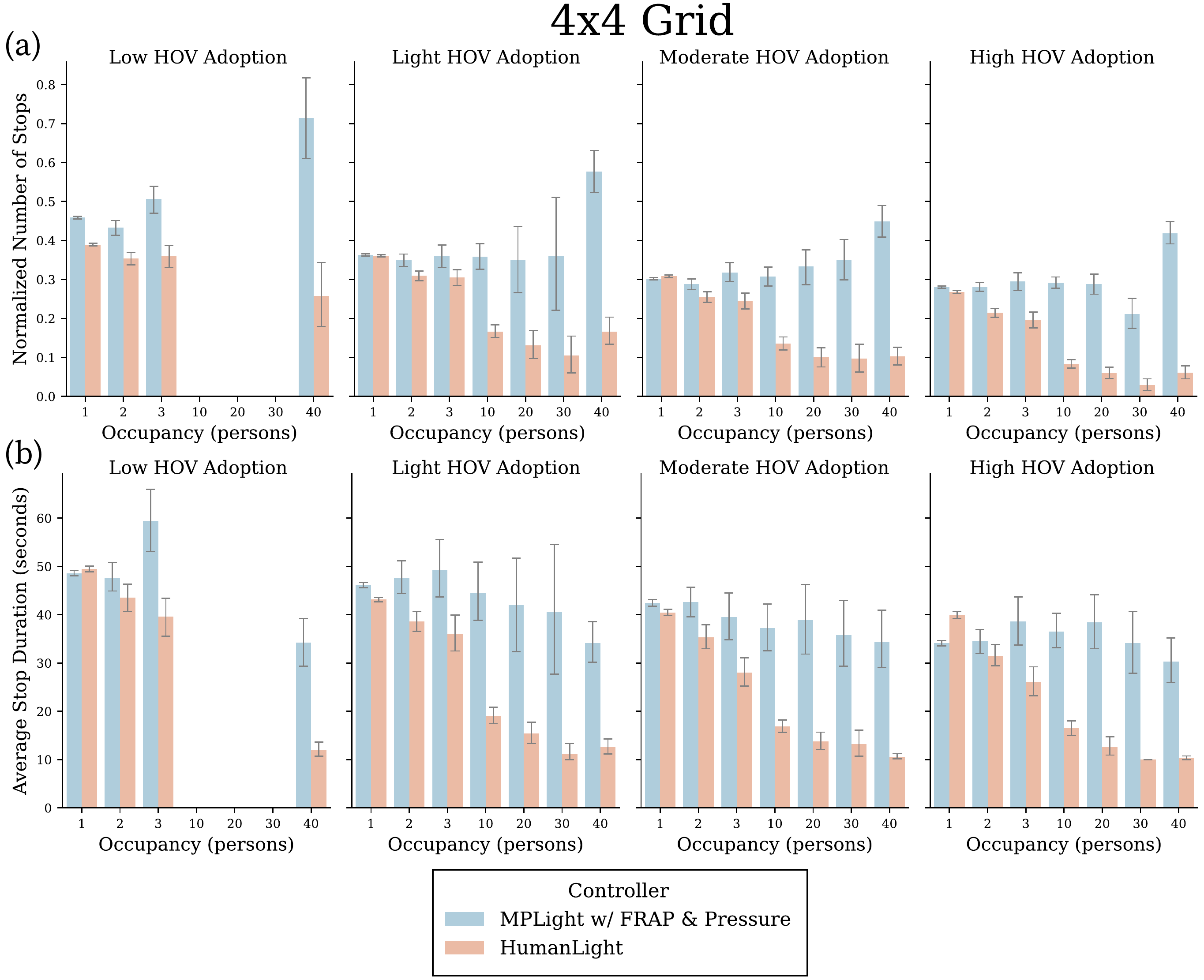}
\caption{Vehicle stopping behavior: Both in terms of normalized per segment stopping frequency (a) as well as average stop duration (b), HumanLight achieves reduced and shorter stops for HOVs as opposed to the vehicle-level approach (MPLight) that generates similar stopping patterns for all vehicles.}
\label{fig:4x4_stop_analytics}
\end{figure}

An analysis on vehicle stopping behavior is carried out in the 4x4 grid network. As mentioned in the queue definition, a vehicle is considered stopped or in queue during the duration it's speed is less than or equal to $0.1 m/s$. We present the per segment normalized number of stops for the various vehicle types to account for the different lengths of routes taken. In Fig. \ref{fig:4x4_stop_analytics}, both in terms of normalized per segment stopping frequency as well as average stop duration, HumanLight achieves reduced and shorter stops for HOVs. Instead, the vehicle-level approach (MPLight) generates similar stopping patterns across vehicle types. The peaks in the normalized stop frequency for transit can be explained from their routeline. As shown in Fig. \ref{fig:roadnets}, bus-lines in the grid scenario never take a right turn. In contrast, all other vehicles' routes may contain right turns that use dedicated right lanes, independent of the current phase and yielding priority only to through moving vehicles in cases of conflict. 

\begin{figure}[ht!]
\centering
\includegraphics[width=1.0\linewidth]{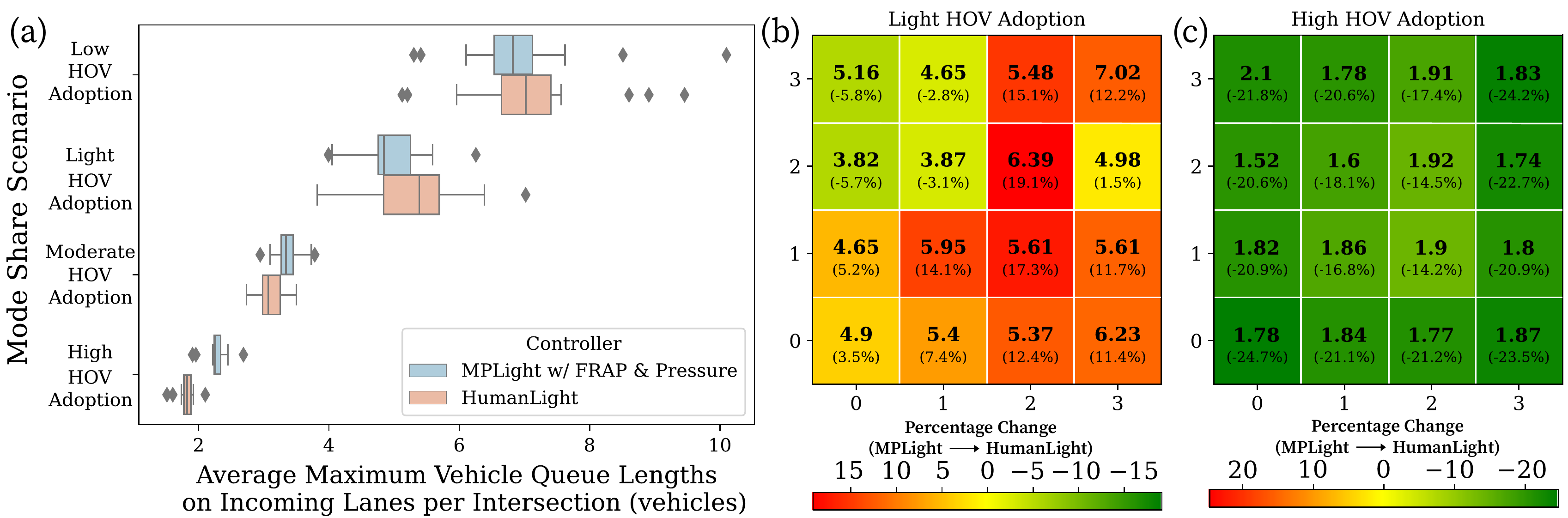}
\caption{(a) Distributions of the maximum queue lengths observed on the incoming lanes for all intersections at every action interval on the 4x4 grid network as generated by MPLight and HumanLight. (b),(c) Maximum vehicle queues observed for the light and high HOV adoption scenarios respectively: In the light scenarios, maximum queues deteriorate on intersections serving the heaviest public transit load, while in the high scenario they improve throughout. Per intersection maximum queues for the other two scenarios are provided in Fig. A\ref{fig:4x4_queues_low_med_hig}.}
\label{fig:4x4_queues_boxplot_light}
\end{figure}

A crucial aspect of implementing a human-centric approach, such as HumanLight, is the impact on vehicle-level metrics. Prioritizing HOVs can lead to network failure by generating queues in non-prioritized directions of traffic. Spill-back effects are caused when vehicle queues exceed road capacities, significantly deteriorating the operational efficiency of the signalized intersections. To quantify that effect on the 4x4 grid network, we evaluate the maximum queue length observed on the incoming lanes of each controlled intersection at every action interval. Fig. \ref{fig:4x4_queues_boxplot_light}a compares the distributions of the averaged over time maximum queues generated from HumanLight versus MPLight, the state of the art vehicle-level approach. Thanks to the consideration of active vehicles in the formulation of the RL problem, HumanLight overall achieves similar or even better maximum vehicle queues while still prioritizing the maximization of people throughput. Only in the low and light HOV adoption scenarios, maximum vehicle queues degrade at some intersections compared to MPLight. 

Fig. \ref{fig:4x4_queues_boxplot_light}b isolates the average, over time, value of the maximum vehicle queues generated by HumanLight for each intersection in the light HOV scenario. In parentheses, the percentage change from MPLight is displayed. We highlight that MPLight is a SOTA vehicle-level controller, not currently in practice. The worst performing intersections experience the heaviest public transit load by serving conflicting and left turning bus routelines. To accommodate the arrival of buses, the controller dictates more frequent phase changes, overwhelming the intersection with stopped vehicles in the non-served traffic movements. Even though for some intersections the maximum queues observed degrade compared to MPLight, the absolute values (illustrated in bold) are still low and far lower from the segments' capacity at jam density. For higher HOV adoption scenarios, the active vehicles consideration in HumanLight renders even better comparative results because of the increased penetration of microtransit and buses (Fig. \ref{fig:4x4_queues_boxplot_light}c). 

\subsection{Active Vehicles: State Space and Reward Formulation Evaluation}
\label{sse_section}

This section evaluates the efficacy of active vehicles (AVs) in the state space and reward function formulation. Intuitively, AVs are conceptualized to provide a more accurate representation of the traffic dynamics. Accounting for vehicles and people not in range of the intersection within the time horizon of the action interval may potentially degrade the performance of control algorithms. Although vehicle counts across the entirety of lanes are extensively used in literature to create embeddings describing the traffic conditions of road segments, passenger loads can significantly fluctuate across vehicles, especially the more the world shifts towards a multimodal reality. These variations can make a substantial difference in the controller's decision making process when determining which phase needs to be prioritized on a person-level approach (Fig \ref{fig:AV_example}). To demonstrate the motivation and the benefits of considering the active vehicles in our person-level approach, we explore three different formulations. In all three alternatives, the current traffic signal phase $\phi^I$ is included in the state space, while the way people counts in the incoming and outgoing lanes are encoded varies as described in Table \ref{tab:av_exploration_scenarios}. All alternatives are illustrated in a corridor of three intersections with the low HOV adoption scenario, while the demand is fixed nearly at capacity as described in Section \ref{sec_experiment}.

\begin{table}[ht!]
\caption{Formulations explored around the efficacy of the active vehicles consideration in the state space and reward function.}
\label{tab:av_exploration_scenarios}
\begin{tabular}{|l|l|l|l|}
\hline
\rowcolor[HTML]{EFEFEF} 
Formulation        & Description                                           & \begin{tabular}[c]{@{}l@{}}State Space \\ Embedding Size\end{tabular} & \begin{tabular}[c]{@{}l@{}}Person Pressure\\ Computed From\end{tabular} \\ \hline
All Vehicles         & Person counts drawn from all vehicles                 & $|L_{in}^I +  L_{out}^I| + 1$                                         & All Vehicles                                                            \\ \hline
Active Vehicles      & Person counts drawn from active vehicles              & $|L_{in}^I +  L_{out}^I| + 1$                                         & Active Vehicles                                                         \\ \hline
All \&Active Vehicle & Person counts drawn from both all and active vehicles & $2\cdot|L_{in}^I +  L_{out}^I| + 1$                                        & Active Vehicles                                                         \\ \hline
\end{tabular}
\end{table}

\begin{figure}[ht!]
\centering
\includegraphics[width=1.0\linewidth]{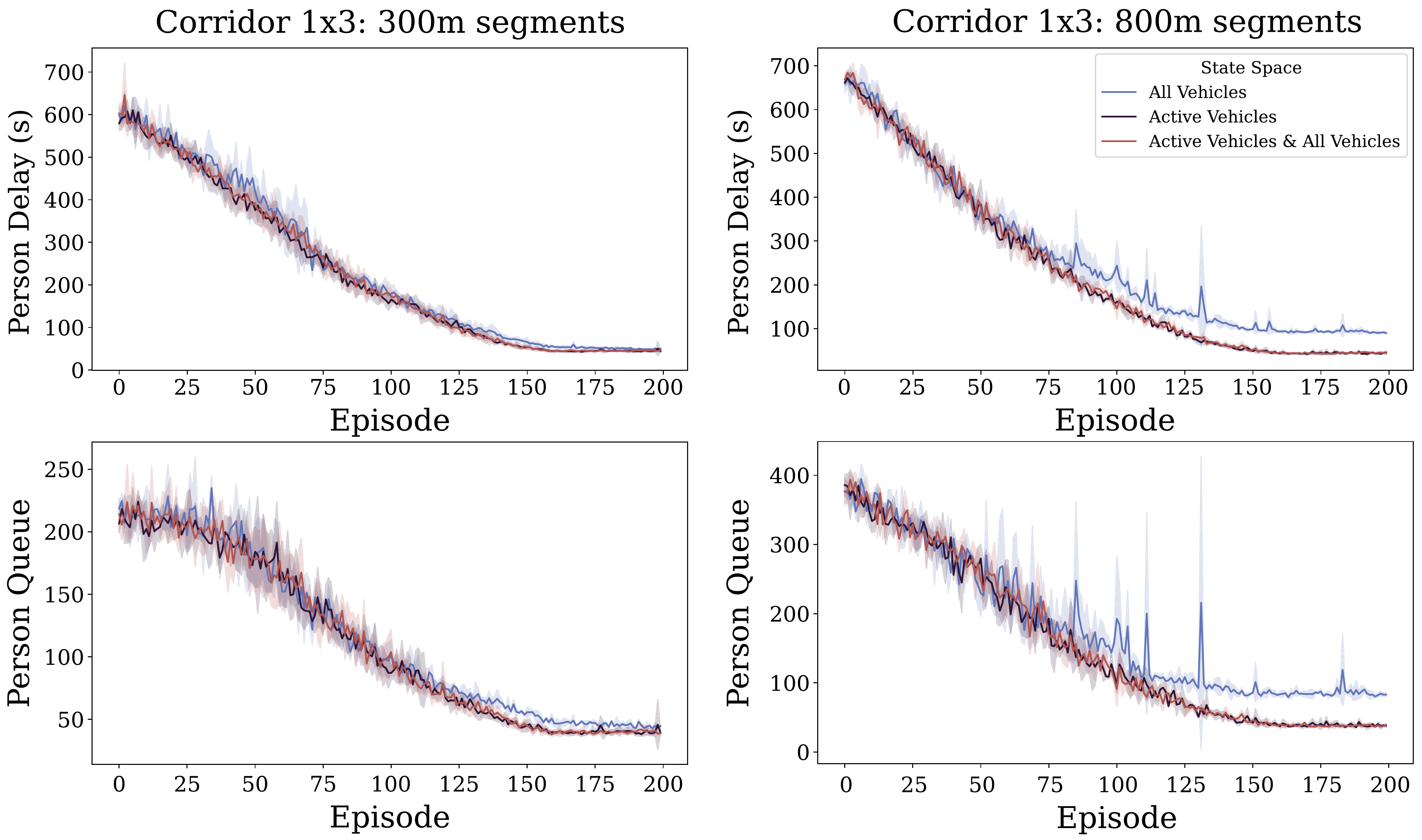}
\caption{Person delays (top) and queues (down) across the three representations of state space for the consideration of active vehicles. Two network structures are tested, one with segments of length 300m (left) and one with segments of length 800m (right).}
\label{fig:sse_plot}
\end{figure}

\begin{table}[ht!]
\caption{Impact of active vehicle consideration on corridors of different segment lengths. Higher benefits are observed for longer road segments.}
\label{tab:AV_impact_table}
\resizebox{\textwidth}{!}{\begin{tabular}{l|llllllll|}
\cline{2-9}
\cellcolor[HTML]{FFFFFF}\textbf{}                                             & \multicolumn{4}{c|}{\cellcolor[HTML]{EFEFEF}\textbf{Delay (s)}}                                                                                                    & \multicolumn{4}{c|}{\cellcolor[HTML]{EFEFEF}\textbf{Queue}}                                                                                      \\ \cline{2-9} 
\cellcolor[HTML]{FFFFFF}\textbf{}                                             & \multicolumn{1}{c|}{Vehicle}       & \multicolumn{1}{c|}{\% Change}         & \multicolumn{1}{c|}{Person}                & \multicolumn{1}{c|}{\% Change}          & \multicolumn{1}{c|}{Vehicle}      & \multicolumn{1}{c|}{\% Change}         & \multicolumn{1}{c|}{Person}        & \multicolumn{1}{c|}{\% Change} \\ \hline
\multicolumn{1}{|l|}{\cellcolor[HTML]{EFEFEF}\textbf{Formulation}} & \multicolumn{8}{c|}{\textit{Corridor 1x3 (300m segments)}}                                                                                                                                                                                                                                                            \\ \hline
\multicolumn{1}{|l|}{All Vehicles}                                            & \multicolumn{1}{l|}{67.99 (2.40)}  & \multicolumn{1}{l|}{Baseline}          & \multicolumn{1}{l|}{49.98 (2.08)}          & \multicolumn{1}{l|}{Baseline}           & \multicolumn{1}{l|}{31.70 (1.06)} & \multicolumn{1}{l|}{Baseline}          & \multicolumn{1}{l|}{45.01 (2.47)}  & Baseline                       \\ \hline
\multicolumn{1}{|l|}{Active Vehicles}                                         & \multicolumn{1}{l|}{61.10 (4.05)}  & \multicolumn{1}{l|}{\textbf{-10.13\%}} & \multicolumn{1}{l|}{\textbf{45.07 (4.24)}} & \multicolumn{1}{l|}{\textbf{-9.82\%}}   & \multicolumn{1}{l|}{28.46 (1.97)} & \multicolumn{1}{l|}{\textbf{-10.22\%}} & \multicolumn{1}{l|}{40.12 (4.87)}  & \textbf{-10.64\%}              \\ \hline
\multicolumn{1}{|l|}{All \& Active Vehicle}                                   & \multicolumn{1}{l|}{61.49 (2.82)}  & \multicolumn{1}{l|}{-9.56\%}           & \multicolumn{1}{l|}{45.25 (2.30)}          & \multicolumn{1}{l|}{-9.46\%}            & \multicolumn{1}{l|}{28.64 (1.42)} & \multicolumn{1}{l|}{-9.65\%}           & \multicolumn{1}{l|}{40.18 (2.51)}  & -10.73\%                       \\ \hline
\multicolumn{1}{|l|}{\cellcolor[HTML]{EFEFEF}\textbf{Formulation}} & \multicolumn{8}{c|}{\textit{Corridor 1x3 (800m segments)}}                                                                                                                                                                                                                                                            \\ \hline
\multicolumn{1}{|l|}{All Vehicles}                                            & \multicolumn{1}{l|}{116.10 (5.69)} & \multicolumn{1}{l|}{Baseline}          & \multicolumn{1}{l|}{93.81 (6.98)}          & \multicolumn{1}{l|}{Baseline}           & \multicolumn{1}{l|}{55.23 (2.51)} & \multicolumn{1}{l|}{Baseline}          & \multicolumn{1}{l|}{87.11 (13.64)} & Baseline                       \\ \hline
\multicolumn{1}{|l|}{Active Vehicles}                                         & \multicolumn{1}{l|}{59.95 (3.27)}  & \multicolumn{1}{l|}{\textbf{-48.36\%}} & \multicolumn{1}{l|}{\textbf{44.35 (2.76)}} & \multicolumn{1}{l|}{\textbf{-52.72 \%}} & \multicolumn{1}{l|}{28.00 (1.69)} & \multicolumn{1}{l|}{\textbf{-49.30\%}} & \multicolumn{1}{l|}{38.67 (3.25)}  & \textbf{-55.61\%}              \\ \hline
\multicolumn{1}{|l|}{All \& Active Vehicle}                                   & \multicolumn{1}{l|}{60.35 (2.23)}  & \multicolumn{1}{l|}{-48.02\%}          & \multicolumn{1}{l|}{44.69 (1.89)}          & \multicolumn{1}{l|}{-52.36\%}           & \multicolumn{1}{l|}{28.05 (1.13)} & \multicolumn{1}{l|}{-49.21\%}          & \multicolumn{1}{l|}{38.16 (2.02)}  & -56.19\%                       \\ \hline
\end{tabular}}
\end{table}

As shown in Eq. \ref{eq:active_range} and Eq. \ref{eq:active_vehicles}, a vehicle being considered as active depends on their relative location on the road segment, their current speed and maximum speed and acceleration. For the set values of vehicle features (Table \ref{tab:sim_model_params}), the active range for a vehicle leaving the intersection under zero speed is $80.25m$ and under maximum speed is $111.11m$. These translate to approximately $1/3$ of the typical road segment length (300m) used in our experiments to be within the active vehicle range. To assess the sensitivity of AVs to the input parameters, we perform the three evaluation alternatives on an additional network structure with longer segments (800m). We observe that all formulations converge after approximately 160 episodes (Fig. \ref{fig:sse_plot}). When considering person counts from active vehicles versus all vehicles, in the 300m segment corridor around 10\% decreases in vehicle and person delays and queues are achieved (Table \ref{tab:AV_impact_table}). The percentage improvements of those metrics almost reach or even exceed 50\% in the 800m segment corridor, while the actual values of delays and queues are similar to the 300m segment corridor. Those results are in alignment with the intuitive expectation that the longer the road segments the higher the impact of the consideration of AVs would be. Table \ref{tab:AV_impact_table} illustrates how the consideration of both active and all vehicles in the state space achieves similar performance as the active vehicle representation alone without further benefits to justify the additional complexity in the embedding. The same experiments were conducted in the same 1x3 corridors under lighter traffic with an average intersection $v/c \approx 0.65$ (Table A\ref{tab:AV_impact_table_light}). Results are consistent with Table \ref{tab:AV_impact_table}. We do notice though that the improvements for the corridor with 800m road segments under light traffic flatten out at approximately 20 \%, hinting that the benefits of AVs are higher for heavier traffic demand.

\subsection{Regulating the aggressiveness of HOV prioritization}

Apart from enabling reduced passenger travel times and socially equitable green time allocation at signalized intersections, HumanLight empowers policymakers and traffic engineers to regulate the aggressiveness in the prioritization of the HOV fleet. We achieve this via a modification in the state embedding where vehicle occupancies are encoded. This way, the travel time benefits across vehicle types of different occupancies is fully in the hands of the system operator rather than generated through a black-box approach. 

\begin{figure}[ht!]
\centering
\includegraphics[width=1.00\linewidth]{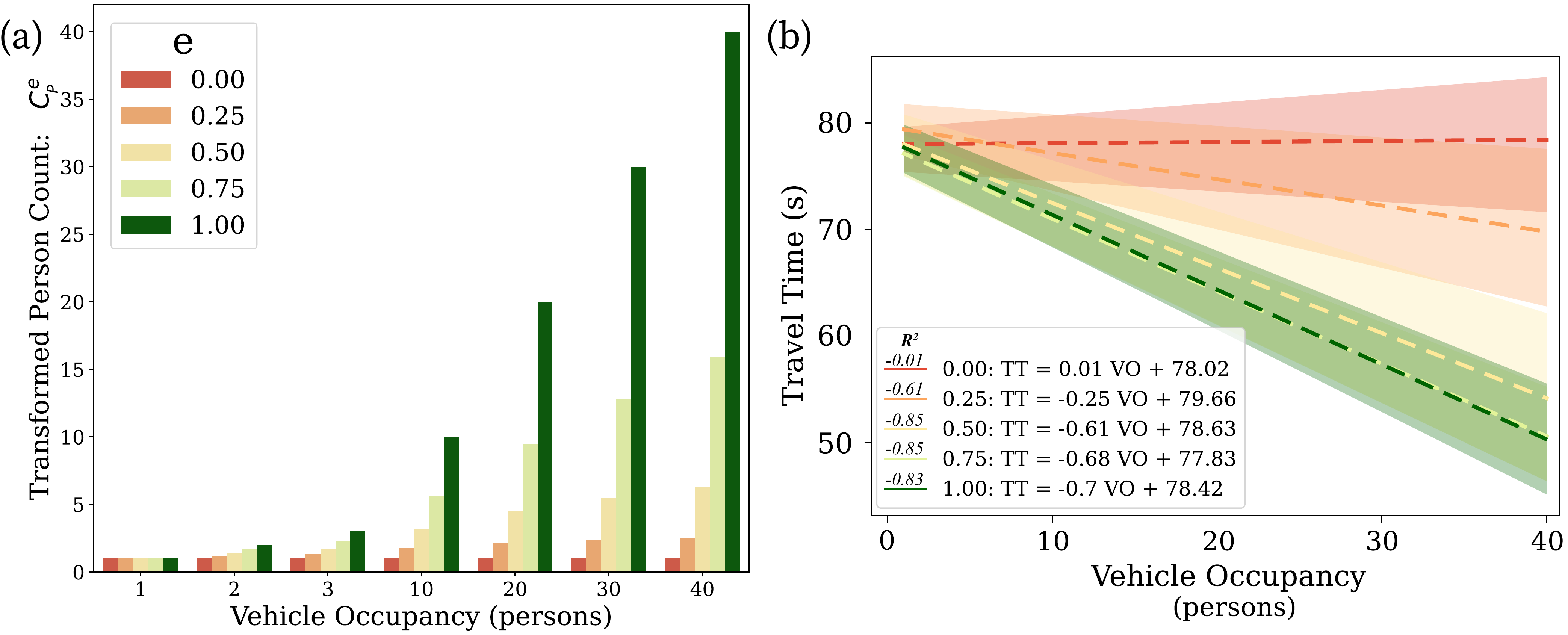}
\caption{(a) Five modified encodings of the person count function for the different vehicle types. (b) Linear least-squares regression fitted  on means of travel times for all vehicle types.} 
\label{fig:prioritization_tuning}
\end{figure}

This concept is illustrated in Figure \ref{fig:prioritization_tuning}a, where five different encodings of person counts are displayed. The dark green representation ($e = 1$) depicts the person count encoding as defined in Section \ref{sec_preliminaries}. With $e$ decreasing, the impact of microtransit and public transit in the reward function compared  to SOVs and carpools is diminishing. On the other extreme, the red representation ($e = 0$) completely cancels out the person count function, turning it into the vehicle counting function, treating all vehicles the same.

We evaluate how HumanLight can regulate the prioritization aggressiveness of vehicles of different occupancies in our high adoption mode share scenario. The high HOV adoption scenario was selected to provide higher number of samples (vehicles) in the microtransit and public transit categories, making the line fitting more robust by definition. Results were reproduced and remain consistent across all mode share scenarios. The evaluation is performed in the single intersection setting. That way every vehicle in our system follows a route of the same length, whether it includes a turn or not, thus making their travel times and delays comparable. Additionally, there is no bias originating from bus pre-specified routes. For example, in the corridor case where buses run along the corridor, vehicles riding from one end to the other without exiting are more likely to get a higher allocation of green times compared to turning vehicles.

Figure \ref{fig:prioritization_tuning}b illustrates how the prioritization of HOVs intensifies with the increase of parameter $e$. A more detailed snapshot of those distributions on the last episode of training can be found in Fig. A\ref{fig:overprioritization_eval_snap}. From evaluating the fitted lines for $e = 0.75$ and $e=1.00$ in Figure \ref{fig:prioritization_tuning}b, we observe that the incline of the travel time vs vehicle occupancy fitted line does not change much in absolute value. That is because HOVs cannot be further over-prioritized. Thus, the two scenarios return similar handling of inter-vehicles conflicts during arrivals at the intersection. As intuitively expected, the fitted line for $e=0$ has an incline of $0.01$, corresponding to a vehicle-level optimization approach. We need to stress that the travel time of $78s$ corresponds to a delay of $24s$ from the minimum free-flow travel time, which is approximately $5s$ less than the vehicle delay of MPLight (Table \ref{tab:si_metrics}) in the same scenario. That improvement is a result of the active vehicle consideration.

Overall, operators of HumanLight should take input from behavioral studies that evaluate the travel time elasticity of commuter mode choice. The aggressiveness in HOV prioritization should be scientifically informed to incentivize riders to mode shift. Over-penalizing SOVs' travel times must not be discouraging to commuters.

\subsection{Impact of Discount Factor $\gamma$ on Phase Profile}
From the hyper-parameters tuned to optimize the performance of our reinforcement learning model, we conducted the most experiments on the discount factor $\gamma$. Apart from a significant impact on performance, the generated policies rendered different phase profiles. This section aims to advise system operators on the impact of prioritising immediate versus long-term rewards. Four values are tested to capture the full spectrum of the effect, ranging from the fully myopic $\gamma = 0$ to $\gamma = 0.90$ that heavily weighs future rewards as described in Eq. \ref{eq:discounted_reward}. Similar to the active vehicle analysis (section \ref{sse_section}), we perform our evaluation on a corridor of three intersections with 300m long segments to grasp any network level effects. We illustrate results under heavy traffic demand (average intersection $v/c$ exceeding 0.92). 

\begin{figure}[ht!]
\centering
\includegraphics[width=0.90\linewidth]{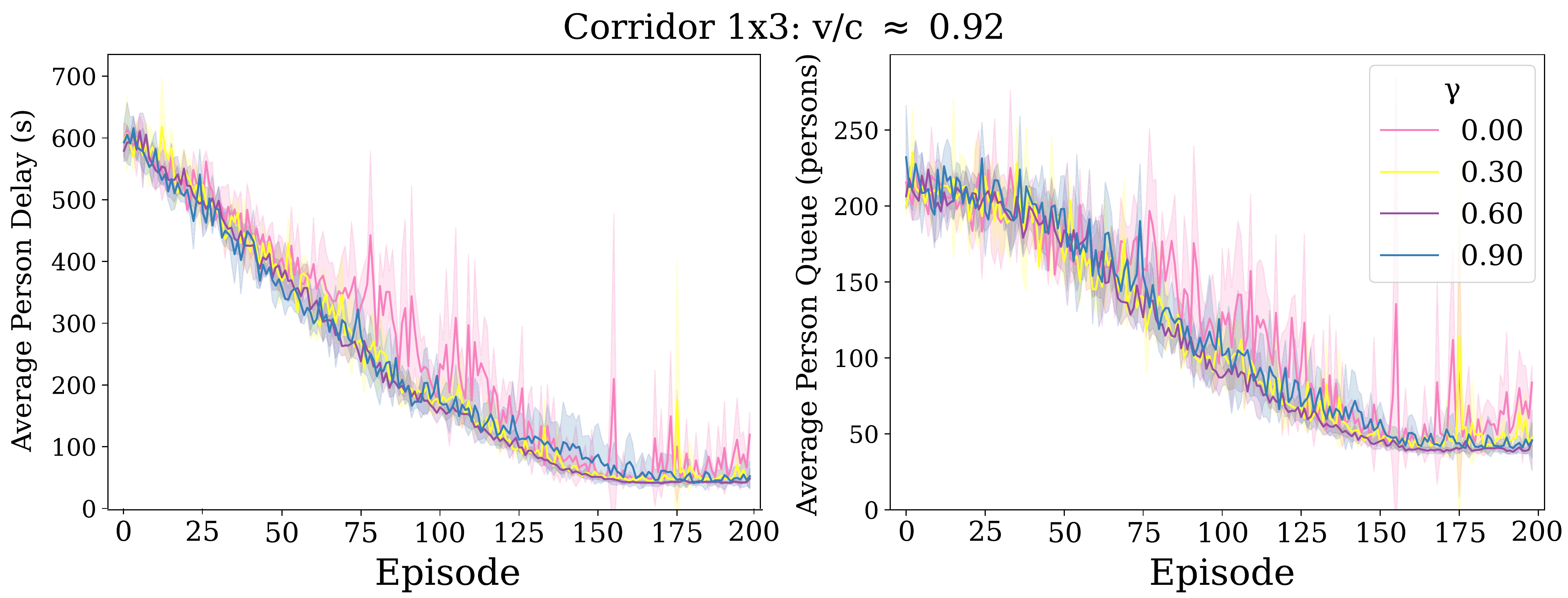}
\caption{Evolution of average person delay and queue for the different values of discount factor $\gamma$ during episodic training for the high traffic demand configuration.}
\label{fig:GammaExploration_convergence_fulldem}
\end{figure}

In terms of convergence, for the lowest $\gamma$ value ($\gamma = 0.00$) we observe significant oscillations even after 200 episodes of training (Fig. \ref{fig:GammaExploration_convergence_fulldem}). Performance-wise, the best performing value for hyper-parameter $\gamma$ is $0.60$ is minimizing delays and queues both in absolute value as well as standard deviation (Table \ref{tab:gamma_impact_table_fulldem}).

\begin{table}[ht!]
\caption{Impact of discount factor $\gamma$ on performance metrics.}
\label{tab:gamma_impact_table_fulldem}
\begin{tabular}{c|cc|cc|}
\cline{2-5}
\multicolumn{1}{l|}{\cellcolor[HTML]{FFFFFF}\textbf{}} & \multicolumn{2}{c|}{\cellcolor[HTML]{EFEFEF}\textbf{Average Delay}} & \multicolumn{2}{c|}{\cellcolor[HTML]{EFEFEF}\textbf{Average Queue}} \\ \hline
\multicolumn{1}{|c|}{\cellcolor[HTML]{EFEFEF}\textbf{$\gamma$}} & \multicolumn{1}{c|}{Vehicle (s)} & Person (s) & \multicolumn{1}{c|}{Vehicle (vehicles)} & Person (persons) \\ \hline
\multicolumn{1}{|c|}{0.00} & \multicolumn{1}{c|}{92.59 (36.45)} & 75.59  (36.18) & \multicolumn{1}{c|}{43.02  (15.25)} & 61.12  (18.71) \\ \hline
\multicolumn{1}{|c|}{0.30} & \multicolumn{1}{c|}{70.52  (11.58)} & 53.41  (10.55) & \multicolumn{1}{c|}{33.07   (5.31)} & 48.15   (9.46) \\ \hline
\multicolumn{1}{|c|}{\textbf{0.60}} & \multicolumn{1}{c|}{61.10 (4.05)} & \textbf{45.07 (4.24)} & \multicolumn{1}{c|}{28.46 (1.97)} & \textbf{40.12 (4.87)} \\ \hline
\multicolumn{1}{|c|}{0.90} & \multicolumn{1}{c|}{65.40  (7.85)} & 51.00  (7.70) & \multicolumn{1}{c|}{31.37  (5.38)} & 44.25  (5.45) \\ \hline
\end{tabular}
\end{table}

\begin{table}[h]
\caption{Sensitivity analysis on the number of phase changes per hour for different values of the discount factor $\gamma$ on two different traffic demand levels at a synthetic corridor 1x3 of intersections.}
\label{tab:phase_changes_vs_gamma_fulldem}
\begin{tabular}{cl|cccc|}
\cline{3-6}
\multicolumn{2}{r|}{}                                                                                                                                             & \multicolumn{4}{c|}{\cellcolor[HTML]{EFEFEF}\textbf{Discount factor $\gamma$}}                                                    \\ \cline{3-6} 
\multicolumn{2}{r|}{\multirow{-2}{*}{\textit{Corridor 1x3 (300m segments)}}}                                                                                      & \multicolumn{1}{c|}{0.00}         & \multicolumn{1}{c|}{0.30}         & \multicolumn{1}{c|}{0.60}          & 0.90          \\ \hline
\multicolumn{1}{|c|}{\cellcolor[HTML]{EFEFEF}}                                                                                                       & v/c  $\approx$ 0.92 & \multicolumn{1}{c|}{56.0 (12.0)}  & \multicolumn{1}{c|}{94.6 (16.3)}  & \multicolumn{1}{c|}{121.9 (10.2)}  & 149.2 (9.8)   \\ \cline{2-6} 
\multicolumn{1}{|c|}{\multirow{-2}{*}{\cellcolor[HTML]{EFEFEF}\textbf{\begin{tabular}[c]{@{}c@{}}Average number of phase \\ changes per hour\end{tabular}}}} & v/c  $\approx$ 0.65 & \multicolumn{1}{c|}{144.3 (28.0)} & \multicolumn{1}{c|}{151.00 (9.1)} & \multicolumn{1}{c|}{151.33 (23.1)} & 175.00 (17.3) \\ \hline
\end{tabular}
\end{table}

\begin{figure}[ht!]
\centering
\includegraphics[width=0.97\linewidth]{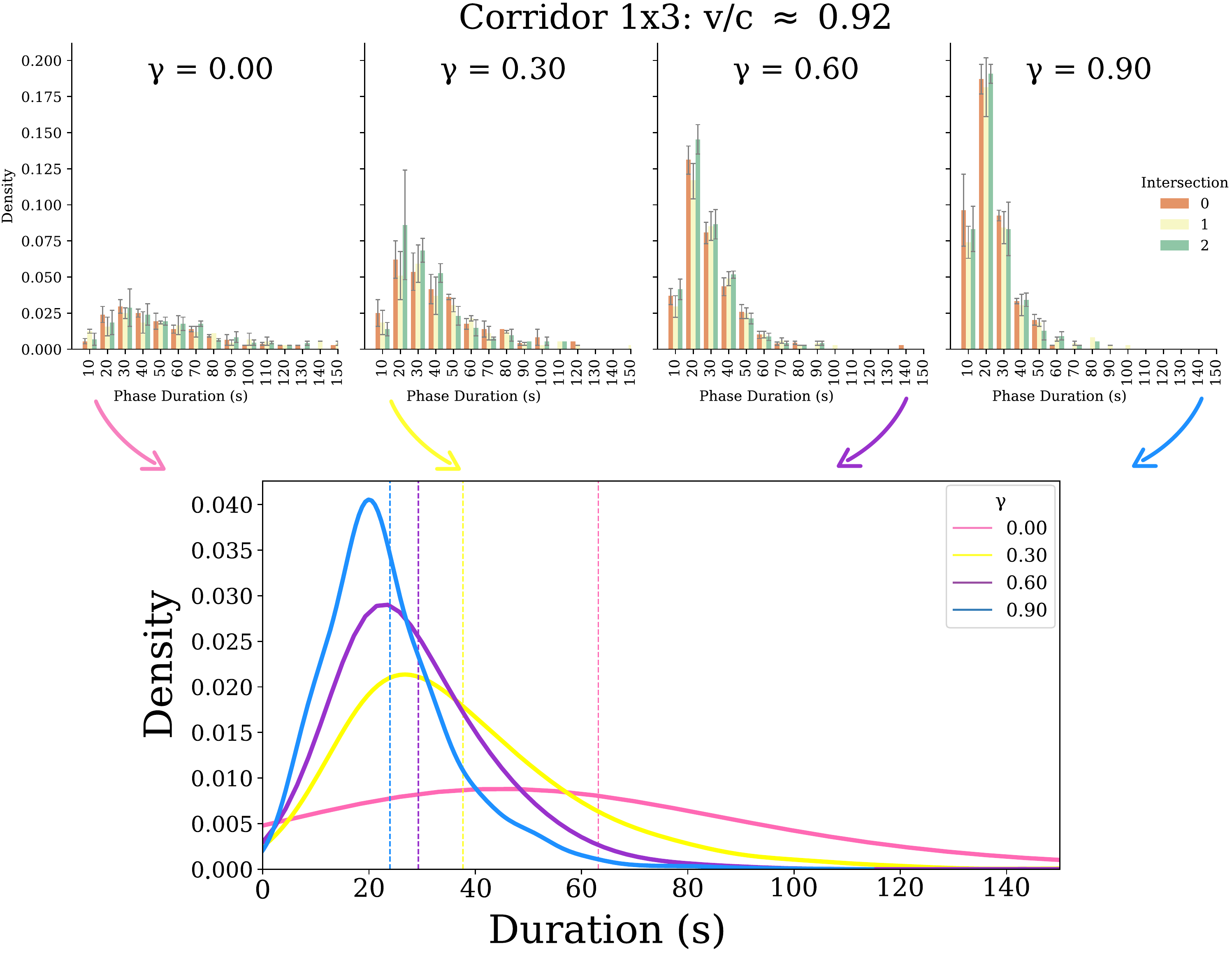}
\caption{Phase profile for different values of $\gamma$ under heavy traffic: A myopic controller, prone to postpone the yellow and all red period will result in less phase changes and higher durations.}
\label{fig:PhaseChangeDur_fulldem}
\end{figure}

When evaluating the impact of the discount factor $\gamma$ to the profile of the generated phases (Table \ref{tab:phase_changes_vs_gamma_fulldem} and Fig. \ref{fig:PhaseChangeDur_fulldem}), we notice that the more short-sighted the controller is (smaller $\gamma$ values), less phase changes are occurring translating to longer durations of the imposed phases. The reason behind this pattern originates from the imposition of the $5 sec.$ yellow and all red period before the initiation of a new (different from the current) phase. A myopic and greedier algorithm will learn to postpone that penalty period where no vehicles are allowed to traverse the intersection of control. Instead, when considering a longer time horizon in the optimization cost function, the controller will be more eager to endure the inevitable penalty period of a phase switch at the appropriate (as dictated by the cost function) point in time. As a benchmark, we note that a typical 8-phase fixed controller with 120 sec. signal length results in $3600/120\cdot8 = 240$ changes per hour. Therefore, the increased number of phase changes (Table \ref{tab:phase_changes_vs_gamma_fulldem}) with the increase of $\gamma$ should not pose any threats to pedestrian crossings waiting time but rather the opposite. Although in this paper HumanLight adopts the value of $\gamma = 0.60$ thanks to its superior performance and balanced phase profile, system operators may chose to sacrifice travel time gains for a desired phase profile. Although HumanLight is built on the acyclic phase scheme, it is fully compatible with a fixed-sequence (cyclic) phase configuration where the algorithm would tune phase extensions and shortenings.

For consistency, the evaluation is also performed under light traffic (average intersection $v/c \approx 0.65$). The results are provided in the appendix (Tables A\ref{tab:gamma_impact_table_070dem} and Fig. A\ref{fig:GammaExploration_convergence_070dem}\&A\ref{fig:PhaseChange&Dur_SM}). Under light traffic, tuning $\gamma$ seems to be less important as $\gamma = 0.60$ and $\gamma = 0.90$ perform equally well. Intuitively, lighter traffic implies the controller will require less consecutive action intervals to fully accommodate the traffic of each direction. This is validated when looking at the number of phase changes per hour (Table \ref{tab:phase_changes_vs_gamma_fulldem}), where we observe that reduced vehicle density leads to increased phase changes. The phase duration profile varies even less than in the heavy traffic case, as long term implications are not as significant.

\section{Conclusions}
\label{sec_conclusions}
This study presents HumanLight, a decentralized adaptive signal control algorithm, which to the best of our knowledge is the first person-based approach founded on reinforcement learning in the context of network level control. Our traffic management solution requires fully developed V2I communication technology to optimize people’s throughput at intersections. HumanLight is designed to equitably allocate green times at intersections. By rewarding HOV commuters with travel time savings for their efforts to merge into a single ride, HumanLight helps democratize urban traffic. 

Considering a mutlimodal vehicle split among private vehicles, pooled rides, microtransit and public transit at various scenarios of HOV adoption, we showcase significant headroom compared to a state-of-the-art RL-based vehicle-level optimization controller. In a 4x4 intersection grid, improvements on person delays and queues exceed 25\% for the lowest HOV adoption scenario which considers no microtransit vehicles and public transit adoption as low as in the COVID-19 pandemic era. For moderate and high HOV adoption, improvements on those same metrics reached over 40\% and 55\% respectively considering a $\sim$26\% and $\sim$41\% shift from single occupancy vehicles to HOV alternatives respectively. As travel time benefits increase along with HOV adoption, HumanLight can invigorate ridesharing and public transit systems to attract the travel demand they truly merit in sustainable and multimodal urban environments.

HumanLight formulates the reward function by extending the transportation theory inspired concept of pressure to the person level. To handle the highly variant occupancy profiles in multimodal urban environments, we introduce active vehicles as vehicles in proximity to the controlled intersections within the action interval window.
Through systematic experiments on varying traffic flows and network structures, we showcase how generating the state space and computing rewards based on active vehicles, enables better decision making for optimizing people’s throughput at intersections.

Apart from rendering reduced passenger travel times and socially equitable green time allocation, HumanLight allows policymakers and traffic engineers to regulate the aggressiveness in the prioritization of the HOV fleet. We illustrate how via a modification in the state embedding, the benefits for the different vehicles types can be tuned, ideally with input from behavioral studies on the travel time elasticity of mode choice. Furthermore, we explore the impact of the discount factor, which determines the importance of future rewards, on performance as well as the generated phase profile. System operators can be informed on the expected phase change and duration patterns which are critical aspects of acyclic signal controllers as they affect pedestrian waiting times. Finally, to assure that HOV prioritization does not lead to excessive queues in the non-prioritized directions of traffic, maximum vehicle queues from all incoming lanes are evaluated in the 4x4 grid setting. Across all scenarios, vehicle queues remain far lower than the segments’ capacity at jam density, posing no threats to the operational efficiency of the signalized intersections.

\section{Future Directions}
\label{future_dirs}
The field of human-centric RL-based traffic signal control is currently under-explored in literature, empowering us to present various directions to be considered for future research. Thanks to the decentralized architecture, HumanLight is a scalable traffic management solution. The algorithmic performance still needs to be evaluated on city-level networks with heterogeneous road and intersection structures. Advancements in high performance computing and simulation \cite{chan2018mobiliti} have made large-scale algorithmic applications possible. As traffic volumes may significantly vary per location, more elaborate designs for coordination and cooperation among neighboring intersections or training multiple controllers may potentially yield better results.

From the demand side, open work contains an exploration at the metropolitan level considering advanced matching algorithms for pooling of trips and quantifying the gains in travel time for different distributions of individuals' walking radius flexibility to join a pooled ride, either at it's origin or on route. For those different levels of adoption, it would interesting to assess the benefits of our human-centric controller in terms of emissions. Although results demonstrate potential as single occupancy vehicles are taken off the roads and HOVs are receiving priority, thus performing less stops, a follow up study quantifying emissions would be of great value. The sensitivity in the framework's effectiveness for varying penetration rates of vehicles with connected technology capabilities should also be evaluated.  

From a planning perspective, the interaction of person-based traffic signal control with dedicated bus lanes needs to be investigated for policymakers to identify the optimal allocation of resources and whether these two HOV prioritization strategies should be applied simultaneously. With HumanLight establishing near free-flow travel times for transit vehicles, the dedicated bus lanes can be freed and used as dedicated bike lanes or  pedestrian safe-spaces, paving the way towards a more sustainable urban environment. As opposed to HOV lanes, our solution also achieves democratization of rides independent of the congestion levels, providing people with incentives to pool even at low traffic demand scenarios.

HumanLight allows the flexibility to balance even more complex priorities of mulitmodal traffic. Bike lanes and pseudo pedestrian lanes, modeling sidewalks, can be included in the codification of the transportation network if demands from these travel means were to be available. Apart from prioritizing bikes and pedestrians via higher weights, HumanLight can accommodate emergency vehicles and paratransit. By assigning ambulances and fire trucks extremely high occupancy values in cases of emergency, these vehicles would receive the highest level of prioritization. Although establishing transit adherence to schedule can be embedded in our reward functions, we believe that if the technology is adopted the bus line schedules should be adjusted accordingly to anticipate the near optimal travel time along the bus route resolving all potential issues of punctuality and bus bunching. With the transit network operating with higher efficiency, the schedule could also become denser offering riders an even higher level of service. In a similar manner, HumanLight can assist policymakers of prosperous urban societies to incentivize ridership of paratransit shuttles. Paratransit provides transportation to people whose disabilities prevent them from riding traditional fixed-route transit service. By accordingly adjusting the weights of paratransit vehicles in the space embedding, HumanLight can make paratransit more attractive for people with disabilities who face widespread lack of accessibility to built environments and roads. By prioritizing these vehicles, cities can foster participation and inclusion of all members of society. 

Further research could explore the effectiveness of our method under cyclic traffic signal configurations. Although research has shown that acyclic phase transitions consistently yield better performance than cyclic phase transitions \cite{kanis2021back}, the fixed sequence provides with more reliability for pedestrian serving across all directions. Other approaches worth exploring include the incorporation of pedestrian waiting time in the reward function where elongated waiting times would be penalized, or the development of an acyclic controller with a minimum green time constraint for some traffic movements. Finally, the impact of the action interval on the benefits from the consideration of active vehicles needs to be investigated. A larger interval will increase the active range across all vehicles, and therefore more vehicles will be accounted for in the state representation.

\section*{Code and Data Availability}
The code, data and relevant documentation are available on Github at \url{https://github.com/DimVlachogiannis/HumanLight.git}.

\section*{Funding}
This work was sponsored by the U.S. Department of Energy (DOE) Vehicle Technologies Office (VTO) under the Big Data Solutions for Mobility Program, an initiative of the Energy Efficient Mobility Systems (EEMS) Program. The following DOE Office of Energy Efficiency and Renewable Energy (EERE) manager played important roles in establishing the project concept, advancing implementation, and providing ongoing guidance: Prasad Gupte. This scientific paper was additionally supported by the Onassis Foundation—Scholarship ID: F ZQ-010/2020–2021.

\section*{Conflict of Interest}
The authors have no known competing financial interests or relationships influencing the work reported to declare.

\section*{Acknowledgements}
We sincerely thank Professor Alex Skabardonis and Professor Daniel Rodriguez for their insightful comments and support.

\printcredits

\clearpage
\appendix
\section*{Appendix}
\refstepcounter{section}
\renewcommand\tablename{Table A.}
\renewcommand\thetable{\unskip\arabic{table}}
\renewcommand\figurename{Figure A.}
\renewcommand\thefigure{\unskip\arabic{figure}}
\setcounter{figure}{0}    
\setcounter{table}{0}    

\begin{figure}[ht!]
\centering
\includegraphics[width=1\linewidth]{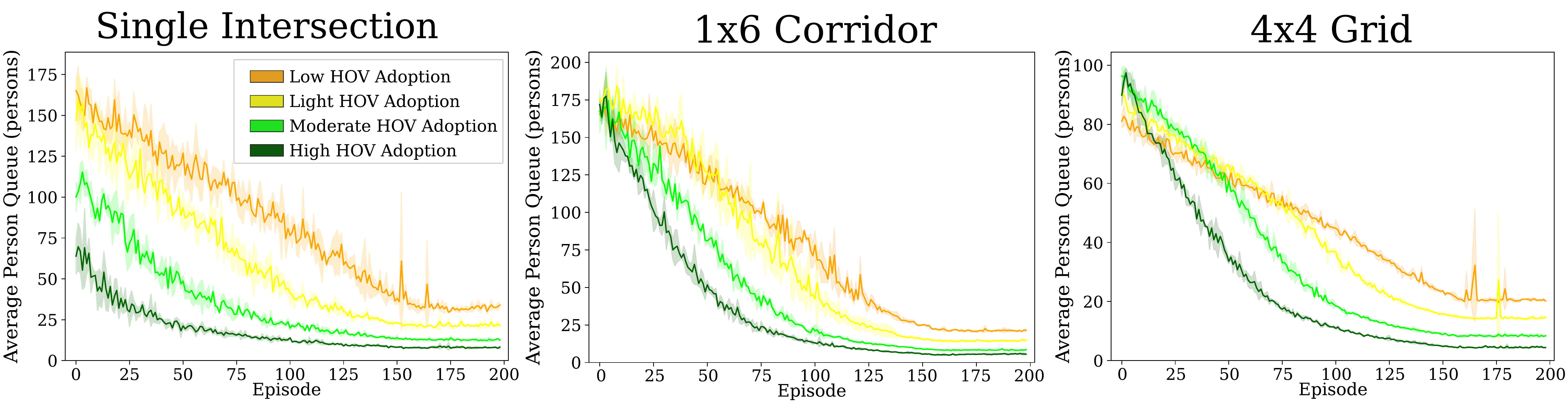}
\caption{Convergence trend of average person queues with HumanLight throughout the learning episodes across all HOV adoptions scenarios.}
\label{fig:person_queue_convergence}
\end{figure}


\begin{table}[h]
\caption{Metric evaluation across all scenarios in the single intersection set of experiments.}
\label{tab:si_metrics}
\resizebox{0.8\textwidth}{!}{\begin{tabular}{cccccc}
\textbf{\begin{tabular}[c]{@{}c@{}}Mode Share \\ Scenario\end{tabular}} & \textbf{Controller} & \textbf{\begin{tabular}[c]{@{}c@{}}Average \\ Vehicle Delay \\ (s)\end{tabular}} & \textbf{\begin{tabular}[c]{@{}c@{}}Average \\ Person Delay \\ (s)\end{tabular}} & \textbf{\begin{tabular}[c]{@{}c@{}}Average \\ Vehicle Queue\\ (vehicles)\end{tabular}} & \textbf{\begin{tabular}[c]{@{}c@{}}Average \\ Person Queue\\ (persons)\end{tabular}} \\ \hline \hline
\multirow{6}{*}{\textbf{{\begin{tabular}[c]{@{}c@{}}Low HOV \\ Adoption\end{tabular}}}} & Fixed (Websters) & 235.76 & 231.86 & 98.09 & 105.75 \\
 & SOTL & 93.38 & 95.11 & 46.53 & 52.23 \\
 & MaxPressure & 62.56 & 62.36 & 31.09 & 33.90 \\
 & MPLight with FRAP + Pressure & 58.75 (2.05) & 59.55 (2.29) & 29.20 (1.14) & 32.47 (1.34) \\
 & HumanLight & 61.25 (3.21) & 58.95 (2.29) & 30.56 (1.78) & 32.11 (1.41) \\
 & \% Improvement from SOTA & \textbf{-4.26\%} & \textbf{1.01\%} & \textbf{-4.66\%} & \textbf{1.11\%} \\ \hline
\multirow{6}{*}{\textbf{\begin{tabular}[c]{@{}c@{}}Little HOV \\ Adoption\end{tabular}}} & Fixed (Websters) & 169.65 & 174.23 & 70.20 & 90.20 \\
 & SOTL & 68.63 & 69.70 & 29.40 & 38.25 \\
 & MaxPressure & 63.36 & 64.01 & 31.15 & 39.98 \\
 & MPLight with FRAP + Pressure & 47.03 (2.63) & 46.68 (2.41)) & 20.16 (1.78) & 25.54 (1.39) \\
 & HumanLight & 47.95 (3.02) & 40.53 (2.56) & 20.54 (1.63) & 21.77 (1.27) \\
 & \% Improvement from SOTA & \textbf{-1.96\%} & \textbf{13.17\%} & \textbf{-1.88\%} & \textbf{14.76\%} \\ \hline
\multirow{6}{*}{\textbf{\begin{tabular}[c]{@{}c@{}}Moderate HOV\\ Adoption\end{tabular}}} & Fixed (Websters) & 109.07 & 116.51 & 40.36 & 67.14 \\
 & SOTL & 54.86 & 52.86 & 18.52 & 27.75 \\
 & MaxPressure & 56.71 & 52.83 & 23.74 & 34.20 \\
 & MPLight with FRAP + Pressure & 38.48 (0.93) & 40.10 (2.37) & 13.05 (0.34) & 21.33 (1.41)\\
 & HumanLight & 35.36 (1.20) & 25.44 (1.03) & 11.82 (0.45) & 12.63 (0.53) \\
 & \% Improvement from SOTA & \textbf{8.11\%} & \textbf{36.56\%} & \textbf{9.43\%} & \textbf{40.79\%} \\ \hline
\multirow{6}{*}{\textbf{\begin{tabular}[c]{@{}c@{}}High HOV\\  Adoption\end{tabular}}} & Fixed (Websters) & 64.64 & 61.43 & 18.17 & 35.59 \\
 & SOTL & 39.62 & 36.50 & 9.81 & 18.67 \\
 & MaxPressure & 53.05 & 57.76 & 17.51 & 39.59 \\
 & MPLight with FRAP + Pressure & 29.04 (0.95) & 28.89 (1.58) & 7.23 (0.26) & 14.88 (1.01) \\
 & HumanLight & 29.31 (1.23) & 16.56 (0.77) & 7.45 (0.38) & 7.96 (0.49) \\
 & \% Improvement from SOTA & \textbf{-0.93\%} & \textbf{42.68\%} & \textbf{-3.04\%} & \textbf{46.51\%} \\ \hline
\end{tabular}}

\end{table}

\begin{table}[ht!]
\caption{Metric evaluation across all scenarios in the 1x6 corridor set of experiments.}
\label{tab:1x6_metrics}
\resizebox{0.85\textwidth}{!}{\begin{tabular}{cccccc}
\textbf{\begin{tabular}[c]{@{}c@{}}Mode Share \\ Scenario\end{tabular}} & \textbf{Controller} & \textbf{\begin{tabular}[c]{@{}c@{}}Average \\ Vehicle Delay \\ (s)\end{tabular}} & \textbf{\begin{tabular}[c]{@{}c@{}}Average \\ Person Delay \\ (s)\end{tabular}} & \textbf{\begin{tabular}[c]{@{}c@{}}Average \\ Vehicle Queue\\ (vehicles)\end{tabular}} & \textbf{\begin{tabular}[c]{@{}c@{}}Average \\ Person Queue\\ (persons)\end{tabular}} \\ \hline \hline
\multirow{6}{*}{\textbf{\begin{tabular}[c]{@{}c@{}}Low HOV \\ Adoption\end{tabular}}} & Fixed (Websters) & 352.49 & 368.53 & 84.89 & 100.16 \\
 & SOTL & 314.97 & 337.22 & 101.53 & 119.69 \\
 & MaxPressure & 68.83 & 73.84 & 26.23 & 30.73 \\
 & MPLight with FRAP + Pressure & 70.22 (4.18) & 73.41 (4.59) & 25.73 (1.92) & 27.82 (1.87) \\
 & HumanLight & 54.78 (3.45) & 53.79 (3.62) & 19.81 (1.20) & 21.24 (1.85) \\
 & \% Improvement from SOTA & \textbf{20.41\%} & \textbf{26.73\%} & \textbf{23.01\%} & \textbf{23.65\%} \\ \hline
\multirow{6}{*}{\textbf{\begin{tabular}[c]{@{}c@{}}Little HOV \\ Adoption\end{tabular}}} & Fixed (Websters) & 237.16 & 268.08 & 63.84 & 92.93 \\
 & SOTL & 120.18 & 153.51 & 38.70 & 63.32 \\
 & MaxPressure & 66.96 & 76.77 & 24.78 & 36.23 \\
 & MPLight with FRAP + Pressure & 48.23 (1.77) & 52.43 (1.88) & 15.67 (0.62) & 21.35 (1.43) \\
 & HumanLight & 41.16 (1.80) & 37.37 (1.82) & 12.95 (0.66) & 14.52 (0.79) \\
 & \% Improvement from SOTA & \textbf{14.66\%} & \textbf{28.72\%} & \textbf{17.36\%} & \textbf{31.99\%} \\ \hline
\multirow{6}{*}{\textbf{\begin{tabular}[c]{@{}c@{}}Moderate HOV \\ Adoption\end{tabular}}} & Fixed (Websters) & 116.06 & 151.72 & 32.17 & 67.73 \\
 & SOTL & 109.07 & 116.51 & 40.36 & 67.14 \\
 & MaxPressure & 59.95 & 70.59 & 19.07 & 35.01 \\
 & MPLight with FRAP + Pressure & 37.11 (0.75) & 42.28 (1.34) & 9.75 (0.24) & 16.28 (1.02) \\
 & HumanLight & 28.77 (1.41) & 22.92 (1.24) & 7.18 (0.45) & 8.35 (0.54) \\
 & \% Improvement from SOTA & \textbf{22.47\%} & \textbf{45.79\%} & \textbf{26.36\%} & \textbf{48.71\%} \\ \hline
\multirow{6}{*}{\textbf{\begin{tabular}[c]{@{}c@{}}High HOV\\  Adoption\end{tabular}}} & Fixed (Websters) & 76.92 & 117.54 & 17.60 & 54.80 \\
 & SOTL & 67.04 & 100.72 & 13.69 & 41.44 \\
 & MaxPressure & 55.31 & 66.04 & 14.41 & 35.00 \\
 & MPLight with FRAP + Pressure & 28.49 (0.98) & 31.13 (1.24) & 5.70 (0.26) & 12.75 (1.55) \\
 & HumanLight & 23.71 (2.11) & 16.30 (1.26) & 4.61 (0.51) & 5.64 (0.68) \\
 & \% Improvement from SOTA & \textbf{16.78\%} & \textbf{47.64\%} & \textbf{19.12\%} & \textbf{55.76\%} \\\hline
\end{tabular}}
\end{table}

\begin{table}[ht!]
\caption{Metric evaluation across all scenarios in the 4x4 grid traffic network configuration.}
\label{tab:4x4_metrics}
\resizebox{0.85\textwidth}{!}{\begin{tabular}{cccccc} 
\textbf{\begin{tabular}[c]{@{}c@{}}Mode Share \\ Scenario\end{tabular}} & \textbf{Controller} & \textbf{\begin{tabular}[c]{@{}c@{}}Average \\ Vehicle Delay \\ (s)\end{tabular}} & \textbf{\begin{tabular}[c]{@{}c@{}}Average \\ Person Delay \\ (s)\end{tabular}} & \textbf{\begin{tabular}[c]{@{}c@{}}Average \\ Vehicle Queue\\ (vehicles)\end{tabular}} & \textbf{\begin{tabular}[c]{@{}c@{}}Average \\ Person Queue\\ (persons)\end{tabular}} \\  \hline \hline
\multirow{6}{*}{\textbf{\begin{tabular}[c]{@{}c@{}}Low HOV\\ Adoption\end{tabular}}}  & Fixed (Websters) & 500.86 & 512.51 & 55.95 & 62.98 \\
 & SOTL & 446.3 & 462.9 & 69.51 & 79.49 \\
 & MaxPressure & 159.39 & 162.97 & 30.74 & 34.49 \\
 & MPLight with FRAP + Pressure & 143.81 (5.62) & 146.29 (5.63) & 25.25 (0.78) & 27.82 (1.36) \\
 & HumanLight & 110.40 (8.23) & 107.00 (8.20) & 19.40 (1.63) & 20.50 (4.10) \\
 & \% Improvement from SOTA & \textbf{23.23\%} & \textbf{26.86\%} & \textbf{23.17\%} & \textbf{26.31\%} \\  \hline
\multirow{6}{*}{\textbf{\begin{tabular}[c]{@{}c@{}}Little HOV\\ Adoption\end{tabular}}} & Fixed (Websters) & 410.14 & 427.24 & 49.51 & 66.24 \\
 & SOTL & 265.09 & 280.01 & 40.33 & 54.34 \\
 & MaxPressure & 139.92 & 141.57 & 24.94 & 32.24 \\
 & MPLight with FRAP + Pressure & 100.94 (2.14) & 103.23 (2.38) & 15.83 (0.31) & 20.14 (0.98) \\
 & HumanLight & 85.20 (3.08) & 74.80 (2.68) & 13.10 (0.54) & 14.40 (0.92) \\
 & \% Improvement from SOTA & \textbf{15.59\%} & \textbf{27.54\%} & \textbf{17.25\%} & \textbf{28.50\%} \\  \hline
\multirow{6}{*}{\textbf{\begin{tabular}[c]{@{}c@{}}Moderate HOV\\ Adoption\end{tabular}}} & Fixed (Websters) & 239.82 & 265.94 & 31.42 & 54.33 \\
 & SOTL & 179.23 & 197.73 & 22.5 & 38.39 \\
 & MaxPressure & 122.33 & 128.66 & 18.99 & 30.92 \\
 & MPLight with FRAP + Pressure & 76.01 (0.91) & 77.86 (1.48) & 9.81 (0.13) & 15.54 (0.70) \\
 & HumanLight & 59.80 (2.60) & 45.30 (2.03) & 7.50 (0.37) & 8.40 (0.55) \\
 & \% Improvement from SOTA & \textbf{21.33\%} & \textbf{41.82\%} & \textbf{23.55\%} & \textbf{45.95\%} \\  \hline
\multirow{6}{*}{\textbf{\begin{tabular}[c]{@{}c@{}}High HOV\\  Adoption\end{tabular}}} & Fixed (Websters) & 169.11 & 193.41 & 18.26 & 41.88 \\
 & SOTL & 136.35 & 154.19 & 13.41 & 30.12 \\
 & MaxPressure & 111.66 & 113.09 & 14.28 & 28.86 \\
 & MPLight with FRAP + Pressure & 57.44 (1.11) & 59.78 (1.45) & 5.76 (0.13) & 11.58 (0.88) \\
 & HumanLight & 40.50 (2.84) & 25.70 (1.80) & 3.90 (0.31) & 4.50 (0.38) \\
 & \% Improvement from SOTA & \textbf{29.49\%} & \textbf{57.01\%} & \textbf{32.29\%} & \textbf{61.14\%} \\\hline
\end{tabular}}
\end{table}
\begin{figure}[ht!]
\centering
\includegraphics[width=0.7\linewidth]{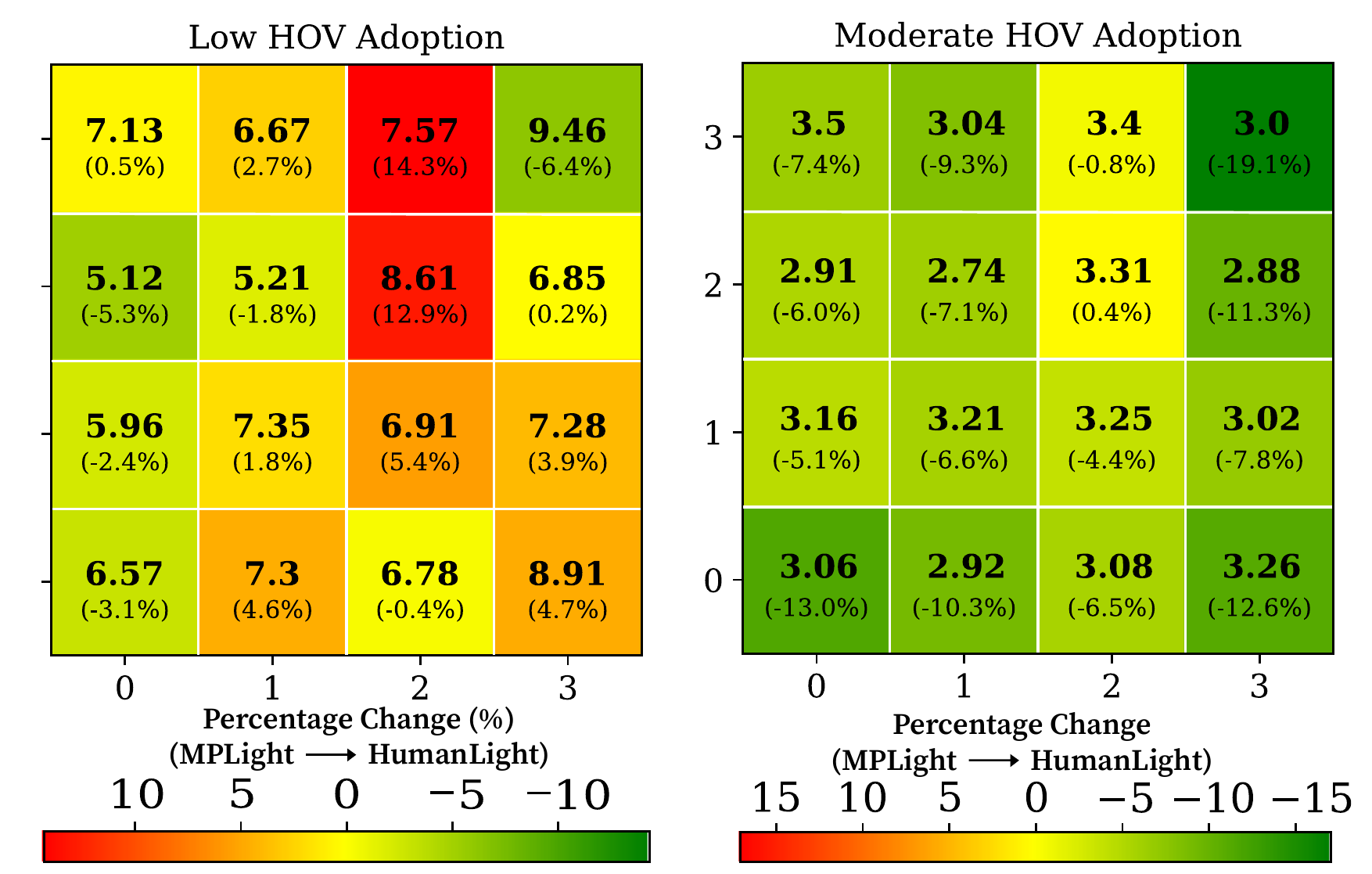}
\caption{Maximum vehicle queue per lane comparison between MPLight and HumanLight: Deteriorations are observed at the low HOV adoption scenario on intersections serving the heaviest public transit load (left). Improvements throughout the grid are reported in the moderate HOV adoption scenario (right).}
\label{fig:4x4_queues_low_med_hig}
\end{figure}

\begin{table}[ht!]
\caption{Impact of active vehicle consideration on experimental corridors under light traffic demand.}
\label{tab:AV_impact_table_light}
\resizebox{\textwidth}{!}{\begin{tabular}{l|llllllll|}
\cline{2-9}
\cellcolor[HTML]{FFFFFF}\textbf{} & \multicolumn{4}{c|}{\cellcolor[HTML]{EFEFEF}\textbf{Delay (s)}} & \multicolumn{4}{c|}{\cellcolor[HTML]{EFEFEF}\textbf{Queue}} \\ \cline{2-9} 
\cellcolor[HTML]{FFFFFF}\textbf{} & \multicolumn{1}{c|}{Vehicle} & \multicolumn{1}{c|}{\% Change} & \multicolumn{1}{c|}{Person} & \multicolumn{1}{c|}{\% Change} & \multicolumn{1}{c|}{Vehicle} & \multicolumn{1}{c|}{\% Change} & \multicolumn{1}{c|}{Person} & \multicolumn{1}{c|}{\% Change} \\ \hline
\multicolumn{1}{|l|}{\cellcolor[HTML]{EFEFEF}\textit{\textbf{Scenario Name}}} & \multicolumn{8}{c|}{\textit{Corridor 1x3 (300m segments)}} \\ \hline
\multicolumn{1}{|l|}{All Vehicles} & \multicolumn{1}{l|}{43.27  (1.38)} & \multicolumn{1}{l|}{Baseline} & \multicolumn{1}{l|}{29.31  (0.86)} & \multicolumn{1}{l|}{Baseline} & \multicolumn{1}{l|}{15.46  (0.55)} & \multicolumn{1}{l|}{Baseline} & \multicolumn{1}{l|}{17.36  (0.58)} & Baseline \\ \hline
\multicolumn{1}{|l|}{Active Vehicles} & \multicolumn{1}{l|}{35.95 (2.09)} & \multicolumn{1}{l|}{\textbf{-16.92\%}} & \multicolumn{1}{l|}{\textbf{24.90  (1.41)}} & \multicolumn{1}{l|}{\textbf{-15.05\%}} & \multicolumn{1}{l|}{12.57  (0.88)} & \multicolumn{1}{l|}{\textbf{-18.69\%}} & \multicolumn{1}{l|}{15.00  (1.18)} & \textbf{-13.59\%} \\ \hline
\multicolumn{1}{|l|}{\cellcolor[HTML]{EFEFEF}\textit{\textbf{Scenario Name}}} & \multicolumn{8}{c|}{\textit{Corridor 1x3 (800m segments)}} \\ \hline
\multicolumn{1}{|l|}{All Vehicles} & \multicolumn{1}{l|}{44.17  (1.72)} & \multicolumn{1}{l|}{Baseline} & \multicolumn{1}{l|}{29.88  (1.27)} & \multicolumn{1}{l|}{Baseline} & \multicolumn{1}{l|}{15.88  (0.69)} & \multicolumn{1}{l|}{Baseline} & \multicolumn{1}{l|}{17.80  (0.84)} & Baseline \\ \hline
\multicolumn{1}{|l|}{Active Vehicles} & \multicolumn{1}{l|}{35.69  (1.86)} & \multicolumn{1}{l|}{\textbf{-19.20\%}} & \multicolumn{1}{l|}{\textbf{24.56  (1.40)}} & \multicolumn{1}{l|}{\textbf{-17.80 \%}} & \multicolumn{1}{l|}{12.48  (0.75)} & \multicolumn{1}{l|}{\textbf{-21.41\%}} & \multicolumn{1}{l|}{14.07  (0.93)} & \textbf{-20.96\%} \\ \hline
\end{tabular}}
\end{table}

\begin{figure}[ht!]
\centering
\includegraphics[width=0.6\linewidth]{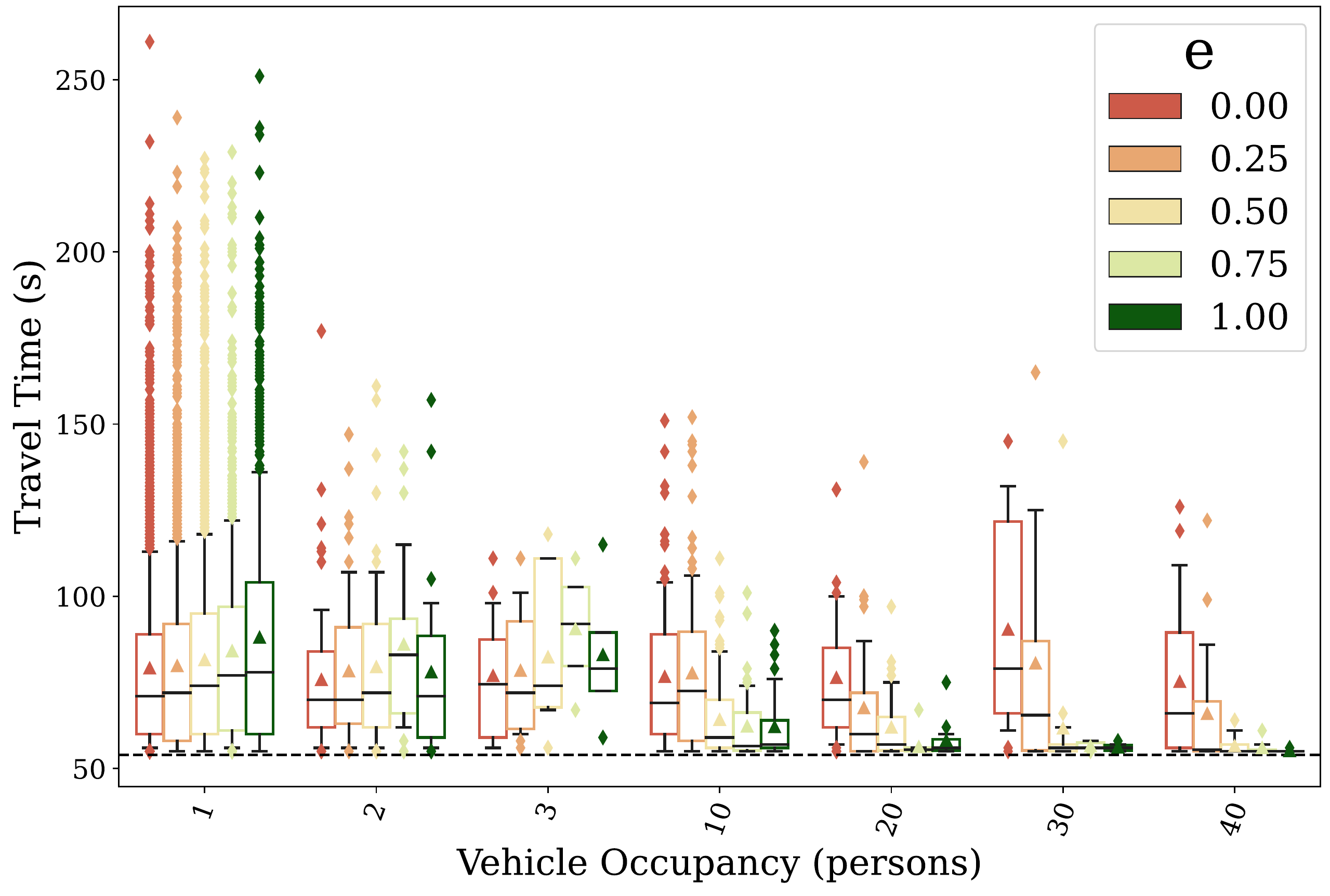}
\caption{For episode 200, the penalization of SOVs in terms of travel time is becoming less intense with the increase of parameter $e$.}
\label{fig:overprioritization_eval_snap}
\end{figure}

\begin{figure}[ht!]
\centering
\includegraphics[width=0.85\linewidth]{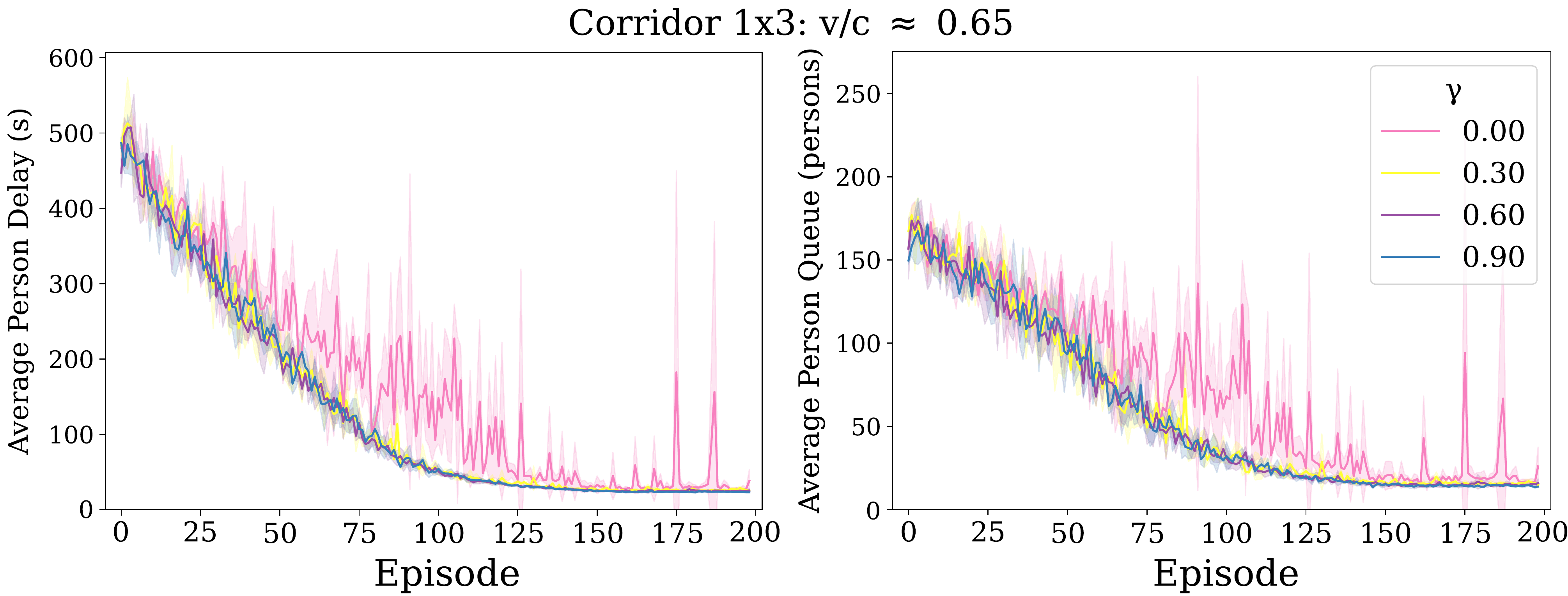}
\caption{Evolution of average person delay and queue for the different values of discount factor $\gamma$ during episodic training for the light traffic demand configuration.}
\label{fig:GammaExploration_convergence_070dem}
\end{figure}
\begin{table}[ht!]
\caption{Impact of discount factor $\gamma$ on performance metrics under light traffic demand.}
\label{tab:gamma_impact_table_070dem}
\begin{tabular}{c|cc|cc|}
\cline{2-5}
\multicolumn{1}{l|}{\cellcolor[HTML]{FFFFFF}\textbf{}} & \multicolumn{2}{c|}{\cellcolor[HTML]{EFEFEF}\textbf{Average Delay}} & \multicolumn{2}{c|}{\cellcolor[HTML]{EFEFEF}\textbf{Average Queue}} \\ \hline
\multicolumn{1}{|c|}{\cellcolor[HTML]{EFEFEF}\textbf{gamma}} & \multicolumn{1}{c|}{Vehicle (s)} & Person (s) & \multicolumn{1}{c|}{Vehicle (vehicles)} & Person (persons) \\ \hline
\multicolumn{1}{|c|}{0.00} & \multicolumn{1}{c|}{50.58  (60.80)} & 39.52  (56.21) & \multicolumn{1}{c|}{17.19  (16.95)} & 22.88  (23.88) \\ \hline
\multicolumn{1}{|c|}{0.30} & \multicolumn{1}{c|}{37.57   (1.80)} & 26.32   (1.69) & \multicolumn{1}{c|}{13.28   (0.72)} & 15.73   (1.17) \\ \hline
\multicolumn{1}{|c|}{\textbf{0.60}} & \multicolumn{1}{c|}{35.95   (2.09)} & \textbf{24.40   (1.41)} & \multicolumn{1}{c|}{12.57   (0.88)} & \textbf{14.70   (1.18)} \\ \hline
\multicolumn{1}{|c|}{\textbf{0.90}} & \multicolumn{1}{c|}{34.01   (0.81)} & \textbf{24.10   (0.59)} & \multicolumn{1}{c|}{11.76   (0.30)} & \textbf{14.52   (0.64)} \\ \hline
\end{tabular}
\end{table}
\begin{figure}[ht!]
\centering
\includegraphics[width=0.70\linewidth]{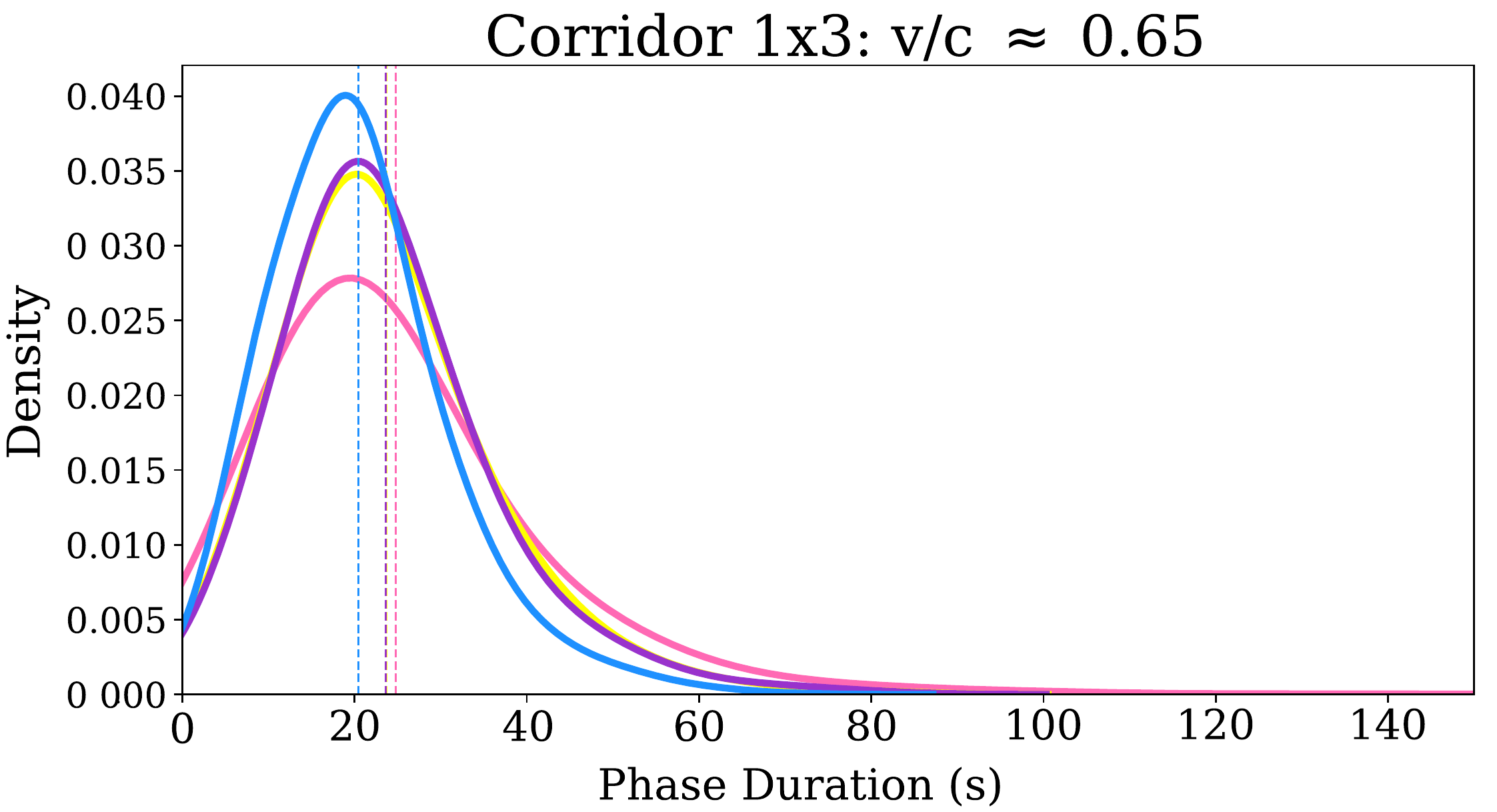}
\caption{Phase profile for different values of $\gamma$ under light traffic.}
\label{fig:PhaseChange&Dur_SM}
\end{figure}

\clearpage
\bibliographystyle{elsarticle-num}

\bibliography{cas-refs}

\end{document}